\pdfoutput=1

\documentclass[11pt]{article}

\usepackage{ACL2023}

\usepackage{times}
\usepackage{latexsym}
\usepackage{comment}

\usepackage[T1]{fontenc}

\usepackage[utf8]{inputenc}

\usepackage{microtype}

\usepackage{inconsolata}

\usepackage{float}

\usepackage{graphicx}
\usepackage{url}
\usepackage{amsmath}
\usepackage[official]{eurosym}

\usepackage{xcolor}
\usepackage{booktabs}
\usepackage{colortbl}
\usepackage{pgf} %
\usepackage{subcaption}
\usepackage{amsmath}
\definecolor{high}{HTML}{ac0026}%
\definecolor{low}{HTML}{fff2ac}%
\newcommand*{\opacity}{60}%
\newcommand*{\minval}{0.30}%
\newcommand*{\maxval}{1.0}%
\newcommand{\gradient}[1]{
    \ifdimcomp{#1pt}{>}{\maxval pt}{#1}{
        \ifdimcomp{#1pt}{<}{\minval pt}{#1}{
            \pgfmathparse{int(round(100*(#1/(\maxval-\minval))-(\minval*(100/(\maxval-\minval)))))}
            \xdef\tempa{\pgfmathresult}
            \cellcolor{high!\tempa!low!\opacity} #1
    }}
}

\usepackage{enumitem}
\definecolor{lightred}{RGB}{255,200,200}

\usepackage{tabularx}

\title{How Humans and LLMs Organize Conceptual Knowledge: \\Exploring Subordinate Categories in Italian}

\author{ \bf
Andrea Pedrotti$^\alpha$, 
Giulia Rambelli$^\beta$,  
Caterina Villani$^\beta$,   
Marianna Bolognesi$^\beta$ \\
$^\alpha$Istituto di Scienza e Tecnologie dell’Informazione “A. Faedo” (ISTI-CNR) \\ \texttt{andrea.pedrotti@isti.cnr.it} \\  $^\beta$Università di Bologna \\ \texttt{\{giulia.rambelli4,caterina.villani6,m.bolognesi\}@unibo.it}
}

\begin{document}
\maketitle
\begin{abstract}
People can categorize the same entity at multiple taxonomic levels, such as basic (\textit{bear}), superordinate (\textit{animal}), and subordinate (\textit{grizzly bear}). While prior research has focused on basic-level categories, this study is the first attempt to examine the organization of categories by analyzing exemplars produced at the subordinate level.
We present a new Italian psycholinguistic dataset of human-generated exemplars for 187 concrete words. We then use these data to evaluate whether textual and vision LLMs produce meaningful exemplars that align with human category organization across three key tasks: exemplar generation, category induction, and typicality judgment.
Our findings show a low alignment between humans and LLMs, consistent with previous studies. However, their performance varies notably across different semantic domains. Ultimately, this study highlights both the promises and the constraints of using AI-generated exemplars to support psychological and linguistic research.\footnote{Data and code is available on \href{https://github.com/andreapdr/conceptCategoriesLLMs}{GitHub} and \href{https://osf.io/875mw/?view_only=21ef4d073c834ed7af46f17e5ae24b11}{OSF}.}

\end{abstract}

\section{Introduction}
\label{sec:introduction}
Concepts are the “building blocks” of human cognition, allowing us to interpret and categorize reality \citep{murphy2004big}.  
The same category can be represented at different levels of inclusiveness (\textit{categorical \textit{specificity}}; \citealp{bolognesi2020abstraction}). For instance, a two-wheeled object may simultaneously be categorized as an \textit{electric bike}, a \textit{bike}, or a \textit{vehicle}, reflecting a hierarchical taxonomy that ranges from a very specific and not inclusive category that only includes members with many common features (\textit{mountain bikes}, \textit{electric bikes}) to a more general and inclusive category that includes a wide variety of items that do not necessarily share many common features (\textit{bike, cars, bus}). %

Most studies on hierarchical organization of categories in the human mind have focused on basic-level categories, showing their advantages in processing and acquisition (\citealp{rosch1976basic,hajibayova2013basic} for a review),
paying little attention to the more specific \textit{subordinate} categories. Yet, words at the subordinate level are crucial for effective communication in specialized domains, as their lexicon conveys richer and more precisely defined semantic content, often derived through linguistic combinations.

\begin{figure}[t!]
    \centering
    \includegraphics[width=0.9\linewidth]{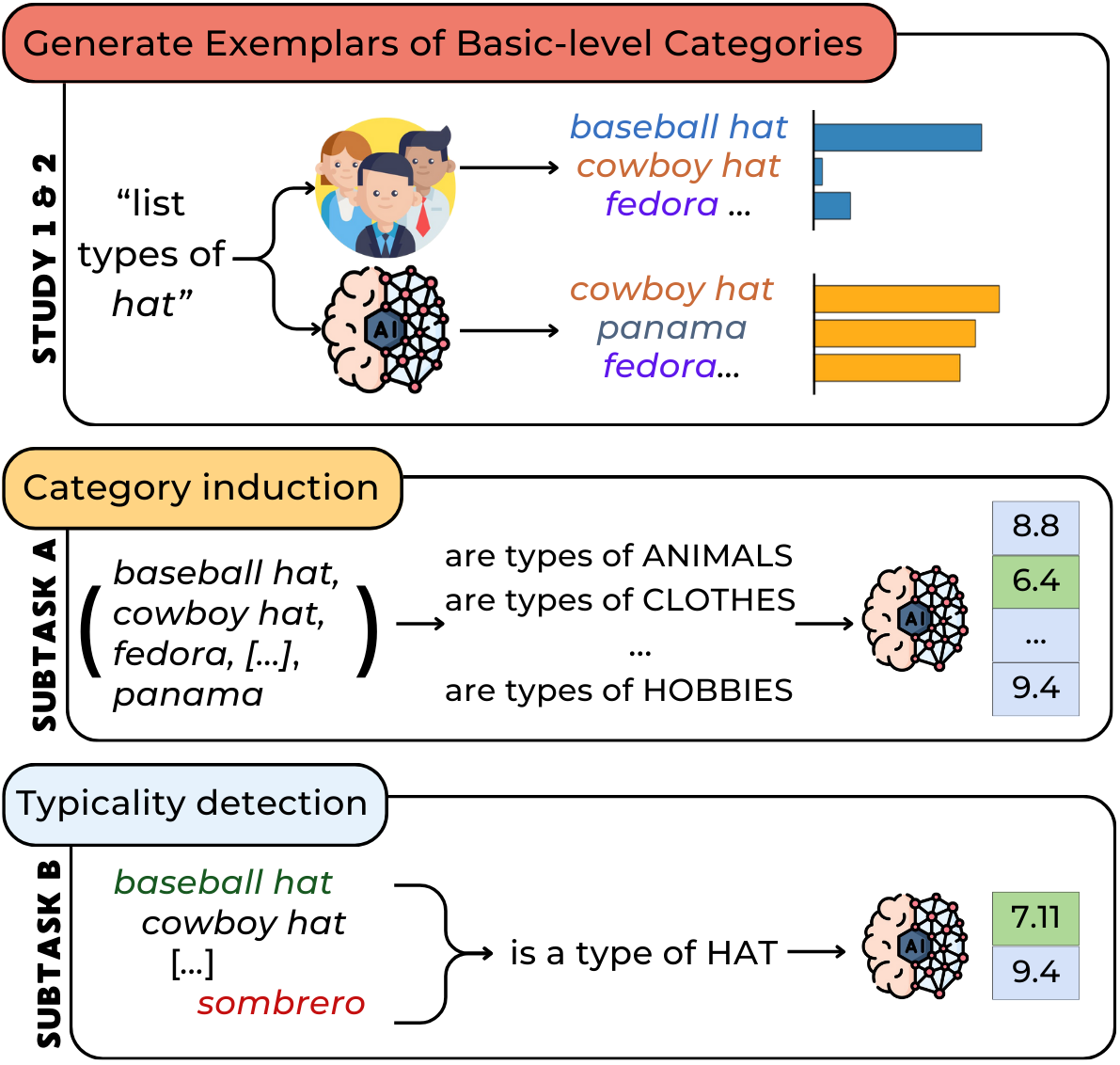}
    \caption{Visual representation of studies’ design. English exemplars are used for illustration only. 
    }
    \label{fig:abstract-image}
\end{figure}

Current cognitive theories acknowledge that both sensorimotor and linguistic experiences contribute to our conceptual representation %
\citep{barsalou2008language,louwerse2018knowing,davis2021building}. For instance, one may observe that \textit{apples} can be red, yellow, or green, but learn in a book that the word \textit{Fuji} refers to a specific variety of apples. Although concepts can be represented independently from words, linguistic labels often act as cues \citep{lupyan2012linguistically,lupyan2019words} that help to create and organize our knowledge, grouping items based on perceived similarities, even if we have never encountered a particular instance before.  The extent to which the organization of human conceptual categories is influenced by the distributional properties of linguistic input remains a central question in cognitive science, linguistics, and artificial intelligence \citep{van2023effects}.

This paper investigates the organization and the contents of conceptual categories produced at a subordinate level by humans and Large Language Models (LLMs). The remarkable success of LLMs raises questions about their plausibility as models of human cognition, 
as their performance closely resembles human-like language understanding and generation across several tasks \citep{wang-etal-2018-glue,brown2020languagemodelsfewshotlearners,floridi2020gpt,bommasani2022opportunitiesrisksfoundationmodels,wei2022emergent}. However, while their functional linguistic competence---reflected in their general knowledge and reasoning skills through language---is undeniable, their parallelism with the human mind remains highly debated (i.a., \citealp{bender-koller-2020-climbing, marcus2020decadeaistepsrobust, MAHOWALD2024517}). 
In contrast to LLMs, human conceptual categories emerge from the integration of linguistic and extra-linguistic (sensory) information.  
 Investigating the structural organization of categories in LLMs may provide insight into the extent to which category formation depends exclusively on linguistic experience; thus, contributing to the larger debate on the role of language in learning semantic knowledge \citep{lupyan2019words}.
While previous works have explored the organization of superordinate categories in both humans and LLMs, we are the first to investigate the organization of basic-level categories. Specifically, we present two studies to address the following research questions: %
\begin{itemize}
    \item \textbf{RQ1: How do humans create and organize basic-level categories, considering the exemplars produced at a subordinate level?}
We introduce a new Italian psycholinguistic dataset, collecting exemplars of 187 basic concrete categories generated by human participants (\S\ref{sec:human-dataset}). We explore the variability of exemplars as a function of category types, assuming that this variability reflects the richness of the linguistic vocabulary and linguistic knowledge in semantic domains. 
\vspace{-0.4em}
\item \textbf{RQ2: Do LLMs have the same category structure as humans?} We probe recent LLMs to generate exemplars for the same 187 basic-level categories and compare their predictions with humans (\S\ref{sec:llms-generation}), as illustrated in Figure \ref{fig:abstract-image}. We assess whether LLMs capture human conceptual organization using two classification subtasks: \textit{category induction} (\S\ref{sec:super-preds}) and \textit{typicality prediction} (\S\ref{sec:typicality}). Finally, we compare vision LMs (vLMs) to investigate whether pre-training extra-linguistic knowledge enhances overall performance.

\end{itemize}

\section{Background and Related Works}
\label{sec:related}
\subsection{Categories in the Human Mind}\label{sec:rel-conceptual}

Classical cognitive research showed that categories are organized hierarchically in the human mind: a \textit{bulldog} is a type of \textit{dog}, which is a type of \textit{mammal}, and more broadly an \textit{animal}, with each category including the previous one.
In other words, categories vary in level of \textit{specificity}---i.e., how inclusive the category of reference is \citep{cohen2005handbook,bolognesi2020abstraction}. 
Superordinate categories (e.g., \textit{furniture, vehicle}) encompass broader classes, while subordinate categories (e.g., \textit{wooden upholstered chairs, red sports cars}) represent more specific instances.  %
The basic level (e.g., \textit{chair, car}), often considered the most informative level, lies between these two extremes, and words that denote basic-level categories are typically easier to understand and process  \citep{murphy2004big}.

A common approach for investigating the structure of categorical knowledge involves analyzing typicality effects, by asking typicality ratings on a Likert scale (i.e., ``How typical is a \textit{cat} for the category \textit{mammal}?'') 
or by instructing participants to freely name members of a given category. The ladder, called ‘‘semantic fluency’’ or ‘‘category instance generation'' \citep{castro2021category}, requires participants to actively retrieve exemplars of a category, which is a more cognitively demanding task than simply judging its typicality within a category. However, typicality ratings can be extracted from category instance generation tasks by aggregating the frequency of exemplar productions. Conversely, words judged as typical for their category are usually more available than words judged to be relatively atypical \cite{HernandezMunoz2006}. In seminal studies, 
\citet{rosch1978principles} observed that some exemplars (e.g., \textit{robin, crow}) are perceived as more representative of a category (e.g., \textit{birds}) than others (e.g., \textit{penguin, ostrich}). This graded structure, as explained by prototype theory \citep{rosch1975cognitive}, reflects the fact that frequently shared properties among category members tend to be integrated into a central prototype.

While cognitive research has extensively focused on superordinate and basic-level categories, subordinate categories have received less attention. Concepts at the subordinate level have some notable peculiarities. First, their referents share more attributes than those within basic-level categories \citep{rosch1976basic}. Additionally, subordinate concepts encode greater perceptual detail, making it more challenging to process individual exemplars. As a result, people tend to name objects at the basic level unless subordinate-level information is particularly relevant. Finally, language plays a crucial role in forming subordinate categories, often created through linguistic compositionality (\textit{electric car, sports car}). To the best of our knowledge, no studies in English or any other language have investigated the organization and contents of basic-level categories (e.g., \textit{dog}, \textit{hammer}), by asking participants to generate concepts at the subordinate level.

\subsection{Categories in LLMs}
\label{subsec:categorystructure-llms}

Previous works on predicting category structure in LLMs have primarily focused on the typicality of a category member, yielding mixed results.
\citet{HeymanHeyman2019} predicted human typicality ratings by correlating similarity scores between category exemplars (e.g., \textit{robin, crow}) and prototype vectors (\textit{bird}), finding that static embeddings poorly accounted for human judgments.  
\citet{renner-etal-2023-exploring} improved predictions using BERT and WordNet metrics, showing that their combination aligns best with human judgments. 
Recently, \citet{heyman2024impact} found out that ChatGPT produces typicality ratings comparable to human participants (.60-.64). 
Conversely, \citet{Misra2021} tested LLMs on taxonomic categorization  (“football is a sport”), showing modest correlations with human ratings (between 0.24 and 0.41) and weaker distinctions between typical and atypical items (as observed in other experimental settings, i.e., \citealp{kauf2023event}). 
Moreover, \citet{misra-etal-2023-comps}  highlighted that LLMs struggle with fine-grained property attributions, questioning their plausibility as models of human semantic memory.

Beyond typicality, \citet{nighojkar2022cognitive} used Transformer models (RoBERTa-Large, DistilBERT, and miniBERTa-med-small) to model the semantic fluency task (\S\ref{sec:rel-conceptual}). They designed different approaches to predict the next item in a given list (“Examples of fruits are the
strawberry and the [MASK]'') for five superordinate categories (Fruits, Vegetables, Animal, Supermarket items, Tool, Foods). Among the models, RoBERTa-Large proved to be the best-performing approach, although it still achieved low performance (16\% overall accuracy).

Concurrently, researchers have investigated whether vision models align with human conceptual understanding \citep{peterson2018evaluating,battleday2020capturing,Gunther2023,upadhyay2022typicality}. Regardless of the specific experimental design, these studies correlated vision-based similarity scores between any pair of exemplar and category images and evaluated these similarities against human typicality judgments.
Recently, \citet{vemuri2024well} evaluated both language and vision models, comparing their correlation with human typicality ratings, and found that textual models are better than vision models for 27 categories, surpassing prior results from \citet{castro2021category}.

Recent works have also tested the abstract reasoning abilities of LLMs. 
For example, \citet{samadarshi2024connecting}  assessed LLMs performance on the \textit{New York Times} Connections game, finding better performance in  Semantic Relations and Encyclopedic Knowledge, which might be due to existing information in pre-training data. However, LLMs accuracy remains below 50\%.

All the aforementioned works focused exclusively on English, and primarily explore the internal organization of superordinate categories (e.g., \textit{fruit, tools}). To our knowledge, no research has yet explored this for Italian or investigated the internal organization of basic-level categories (e.g., \textit{dog, hammer}).

\section{\textsc{Study 1}:  A New Psycholinguistic Dataset of Basic-Level Exemplars}
\label{sec:human-dataset}
\paragraph{Methods.}\label{sec:human-dataset-method}
Stimuli consist of 187 basic-level concrete categories previously produced by Italian native speakers as the most representative concepts for 12 superordinate semantic categories\footnote{ANIMALS, BODY PARTS, CLOTHES, FOODS, FURNISHING, FURNITURE, HOBBIES, HOUSING, KITCHEN, PLANTS, STATIONERY, VEHICLES.} \citep{Montefinese2012SemanticMA}. 
We administer an exemplar generation task to 365 Italian L1 speakers on Prolific. Each participant is presented with a list of 15-16 categories and asked to produce as many exemplars as possible for each concept (e.g., \texttt{List a type of}) at their own pace. The final dataset, after post-processing typos and misspellings, consists of 24.659 exemplars.

We compute the same measures as \citet{Montefinese2012SemanticMA} to describe the 
relationship between a given concept and its exemplars, such as the proportion of participants who produce a target exemplar given a category (\textit{dominance}), the mean output position of each exemplar for a category (\textit{mean rank order}), and the proportion of participants who produce a given exemplar as their first response (\textit{first occurrence value}). We primarily focus on \textit{exemplar availability}, which represents how readily an exemplar is produced as a member of a category. This measure is determined by the exemplar's position in a participant’s response list, its overall production frequency within the category, the earliest position it appears across participants, and the total number of participants who mention it.

\paragraph{Results and Discussion.}  
In line with \citet{Montefinese2012SemanticMA}, we find that dominance, availability, and first occurrence are all strongly and positively correlated (\textit{r}$_s$ = 0.95, 0.75, 0.89; for dominance vs. availability, dominance vs. first occurrence, and availability vs. first occurrence); whereas mean rank order of production correlates weakly and negatively with the other three measures (\textit{r}$_s$ = - 0.09; -0.21, -0.15; for dominance, first occurrence, and availability respectively). To identify the most representative exemplars for each concept, we retain only those exemplars with a dominance value higher than or equal to 0.1 (i.e., exemplars produced by at least 10\% of participants). This cut-off criterion results in a total of \textbf{1696 exemplars} in the final dataset.

\begin{figure}[h!bt]
    \centering
    \includegraphics[width=1.0\linewidth]{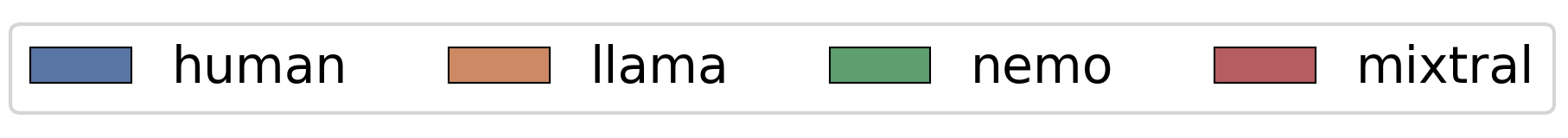}
    \includegraphics[width=1.0\linewidth]{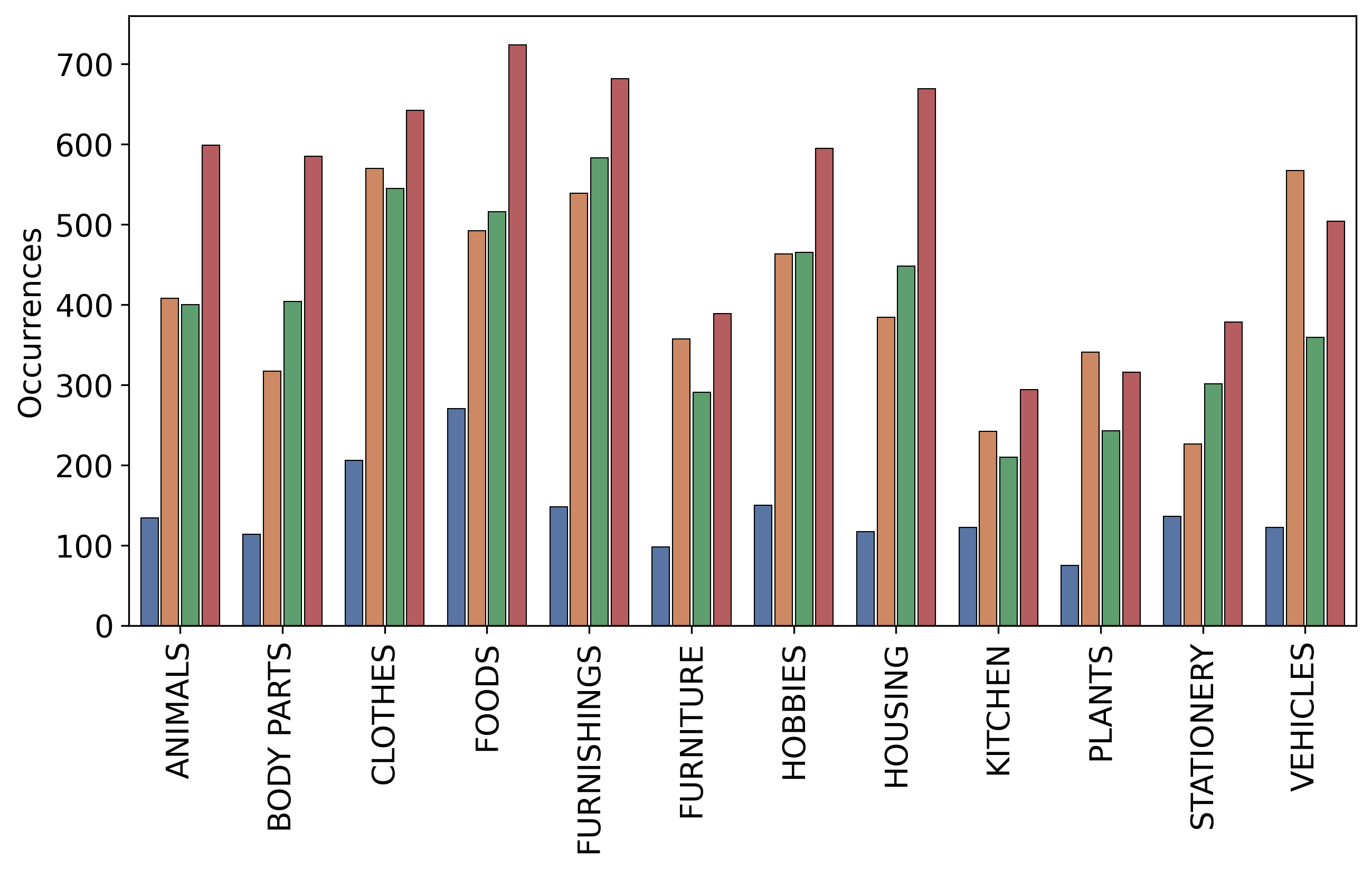}
    \caption{Number of valid exemplars across 12 superordinate categories for humans and textual LLMs.
    }
    \label{fig:human_bycateg}
\end{figure}

Figure \ref{fig:human_bycateg} shows the numbers of dominant exemplars for the 12 subordinate categories. The highest number of exemplars is produced for the \textbf{FOOD} category (\textbf{270} exemplars), followed by \textbf{CLOTHES} (\textbf{206} exemplars), whereas the category of PLANTS has the smallest number of exemplars (77). Indeed, the number of dominant exemplars varies considerably within each basic-level category, spanning from a minimum of 1 exemplar (e.g., \textit{sunflower}, \textit{rubber plant}) to a maximum of 31 exemplars (e.g., \textit{pasta}, \textit{dog}). The fact that the extent of our subordinate lexicon varies in human cognition suggests that \textbf{some subordinated categories might pose challenges in terms of accessibility to semantic memory}. This could be due to their low frequency or familiarity, or to a higher degree of individual variability in knowledge within a specific domain compared to others. 

Subsequently,  for each basic-level category, we compare the top-1 and top-5 exemplars ordered by availability with those ordered by dominance, examining whether both the exemplars and their order align. %
Overall, 77.0\% of the top-1 dominant exemplar is also the top-1 available. This indicates that more frequently produced exemplars tend to be more readily available in participants' responses, reflecting their prominence in the conceptual category. However, only 13.9\% of the top-5 dominant exemplars overlap with the ranking of the top-5 most available exemplars. For example, the top-5 dominant exemplars for the basic category “\textit{cereal}” are \textit{oat, spelt, wheat, corn, barley}, while the top-5 available were \textit{wheat, oat, spelt, corn, barley}. This outcome points to potential variability in the frequency of production and availability of exemplars within different conceptual categories. In conclusion, we observe that this task is more challenging than retrieving exemplars of superordinate-level categories and that some categories are more accessible than others.

\section{\textsc{Study 2}: LLMs' Exemplars Generation}
\label{sec:llms-generation}

\renewcommand*{\minval}{0.3}%
\renewcommand*{\maxval}{1.0}%

\begin{table*}[t!]
\resizebox{\linewidth}{!}{%
    
\begin{tabular}{lccccccccccccr}
\toprule
 & \MakeUppercase{animals} & \MakeUppercase{body parts} & \MakeUppercase{clothes} & \MakeUppercase{foods} & \MakeUppercase{furnishing} & \MakeUppercase{furniture} & \MakeUppercase{hobbies} & \MakeUppercase{housing} & \MakeUppercase{kitchen} & \MakeUppercase{plants} & \MakeUppercase{stationery} & \MakeUppercase{vehicles} & avg \\
\midrule
\texttt{llama-3.2-3B}       & \gradient{0.63} & \gradient{0.74} & \gradient{0.76} & \gradient{0.84} & \gradient{0.61} & \gradient{0.69} & \gradient{0.67} & \gradient{0.72} & \gradient{0.59} & \gradient{0.50} & \gradient{0.63} & \gradient{0.63} & 0.67 \\
\texttt{llama-3.1-8B}       & \gradient{0.48} & \gradient{0.64} & \gradient{0.64} & \gradient{0.86} & \gradient{0.61} & \gradient{0.69} & \gradient{0.74} & \gradient{0.72} & \gradient{0.49} & \gradient{0.41} & \gradient{0.49} & \gradient{0.63} & 0.62 \\
\texttt{llama-3.1-70B}   & \gradient{0.81} & \gradient{0.77} & \gradient{0.89} & \gradient{0.98} & \gradient{0.82} & \gradient{0.83} & \gradient{0.85} & \gradient{0.93} & \gradient{0.80} & \gradient{0.68} & \gradient{0.61} & \gradient{0.83} & \textbf{0.82} \\
\texttt{mistral-7B}         & \gradient{0.52} & \gradient{0.61} & \gradient{0.43} & \gradient{0.79} & \gradient{0.50} & \gradient{0.41} & \gradient{0.69} & \gradient{0.61} & \gradient{0.46} & \gradient{0.42} & \gradient{0.39} & \gradient{0.57} & 0.53 \\
\texttt{nemo-12B}            & \gradient{0.71} & \gradient{0.72} & \gradient{0.69} & \gradient{0.90} & \gradient{0.71} & \gradient{0.65} & \gradient{0.79} & \gradient{0.86} & \gradient{0.56} & \gradient{0.47} & \gradient{0.58} & \gradient{0.70} & 0.69 \\
\texttt{mixtral-8x7B}         & \gradient{0.73} & \gradient{0.76} & \gradient{0.77} & \gradient{0.95} & \gradient{0.74} & \gradient{0.76} & \gradient{0.81} & \gradient{0.86} & \gradient{0.67} & \gradient{0.54} & \gradient{0.53} & \gradient{0.79} & 0.74 \\
\midrule
\texttt{llava-7B}           & \gradient{0.52} & \gradient{0.60} & \gradient{0.54} & \gradient{0.67} & \gradient{0.57} & \gradient{0.53} & \gradient{0.70} & \gradient{0.61} & \gradient{0.48} & \gradient{0.48} & \gradient{0.57} & \gradient{0.61} & 0.57 \\
\texttt{idefics2-8B}        & \gradient{0.64} & \gradient{0.76} & \gradient{0.62} & \gradient{0.80} & \gradient{0.75} & \gradient{0.67} & \gradient{0.82} & \gradient{0.71} & \gradient{0.53} & \gradient{0.67} & \gradient{0.65} & \gradient{0.65} & 0.69 \\

\bottomrule
category avg         & 0.63 & 0.70 & 0.67 & \textbf{0.85} & 0.66 & 0.66 & 0.76 & 0.75 & 0.57 & 0.52 & 0.56 & 0.68 & 0.67 \\
\bottomrule
\end{tabular}
}
\caption{Percentage of valid exemplars generated by various LLMs. 
}
\label{table:validity-bycat}
\end{table*}

We probe several LLMs on the task described in \S \ref{sec:human-dataset} to compare their 
organization of subordinate-level conceptual representations with human subjects. We assess models' performance considering:
(\textit{i}) the number of hallucinations generated (i.e., non-existent exemplars created by combining words into \textit{ad hoc} instances);
(\textit{ii}) the overlap with human subjects regarding the most available (typical) exemplar, and
(\textit{iii}) whether discrepancies between human and LLMs-generated exemplars follow a consistent pattern. 

We analyse our data from two complementary perspectives. On the one hand, we assess the models' accuracy based on their similarity to human-generated exemplars (our gold standard). On the other, we perform some qualitative analyses to explore whether and how the categorical knowledge encoded by language models differs from that of humans.

\paragraph{Setup.} Building upon the methodology described in \S \ref{sec:human-dataset}, we task the models with generating exemplars for the same 187 basic-level concepts presented to human subjects. We use two LLMs families: \textit{(i)} \textbf{LLaMA family}, including \texttt{LLaMA-v3.1} in its 8 and 70B versions, and \texttt{LLaMA-v3.2-3B} \citep{dubey2024llama3herdmodels}, and \textit{(ii)} \textbf{Mistral family}, comprising \texttt{Mistral-7B} \cite{jiang2023mistral7b}, \texttt{Mixtral-8x7B} \cite{Jiang2024MixtralOE}, and \texttt{NeMO}\footnote{\url{https://mistral.ai/news/mistral-nemo/}}.
Furthermore, to investigate the impact of perceptual extra-linguistic stimulus, we also use the vLMs \textbf{LLaVA} \citep{liu2024visual} and \textbf{Idefics2} \citep{laurenccon2024matters} (cf. Appendix \ref{app:models} for an in-depth description).%

We model the generation process as a few-shot setting \citep{brown2020languagemodelsfewshotlearners} completion task. 
The model receives a simplified version of the instructions from \S\ref{sec:human-dataset} to obtain comparable results. The instruction is followed by a question-answer example before generating exemplars for a new concept.
We follow the few-shot prompting scenario, as this approach should positively affect the model’s performance. We experiment with parameters to obtain an outcome balanced between predictability and creativity. For each model, we perform five runs for each basic-level category (cf. Appendix~\ref{app:llm-gen-prompt}). 

\subsection{Analysis 1: LLMs Tends to Generate \textit{ad hoc} Expressions instead of Exemplars}
The generated responses consist of a list of exemplars separated by newlines (i.e., `\texttt{\textbackslash n}'). To ensure data quality, we first clean the outputs by removing duplicate exemplars, keeping only their first occurrences. We then validate the outputs by checking whether each exemplar appears at least once in the Italian corpus \texttt{ItTenTen}
\citep{jakubivcek2013tenten, suchomel2012efficient}\footnote{We use the SketchEngine API to collect frequencies.}, thereby distinguishing valid exemplars from (possible) hallucinations.

This data-cleaning step allows for an overall evaluation of the quality of the generated exemplars in terms of the percentage of valid (i.e., existing expression) exemplars. %
Table~\ref{table:validity-bycat} shows that the performance differs widely across models, with larger and more recent LLMs generating a higher proportion of valid exemplars in comparison to smaller models or vLMs. For instance, \texttt{LLaMA-v3.1-70B} generates 82\% valid exemplars, while \texttt{Mistral-7B} generates only 52\% valid exemplars. The lowest performance is observed by \texttt{LLaVa-7B} (44\%).

Notably, the number of valid exemplars varies depending on the superordinate category. Categories, such as \textbf{FOOD} (\textbf{85\%}), \textbf{HOBBIES} (\textbf{76\%}), and \textbf{HOUSING} (\textbf{75\%}), yield a higher proportion of valid exemplars across models. In contrast, categories like KITCHEN and PLANTS exhibit more noise, with only 57\% and 52\% of valid exemplars, respectively. 
This indicates that models acquire a non-uniform knowledge of subordinate-level exemplars, with a broader and more precise coverage of certain basic-level concepts, while showing a more brittle grasp of others. These results partially align with human behaviour: the categories' exemplars that are easiest (FOOD) and those that are most difficult (PLANTS) to recall are the same for both humans and LLMs.

Considering unattested expressions,  LLMs often rely on their compositional abilities to generate surface-acceptable expressions. However, this `creative' process produces invalid multi-word expressions (i.e., hallucinations) that lack validation among human speakers (i.e., their corpus frequency is zero) and/or real-world referents. 
We conduct a qualitative analysis of zero-frequency items to identify recurring generative tendencies on \texttt{LLaMA-3.1-70B} (the best-performing model in terms of valid exemplars generated). %
Among others, we observe that the model tends to replicate the surface-level syntactic or morphological structure of a valid, attested exemplar, leading to the overgeneralization of that structure to produce novel combinations. For instance, the expression \textit{abete rosso} (`red fir') and \textit{abete di Douglas} (‘Douglas fir’) serve as a template for generating further expressions like \textit{abete bianco di Scozia} (`white Scotch fir') or \textit{abete rosso di California} (`red California fir'), none of which refer to real-world referents. 
Similarly, the models extract from \textit{candelabro a 5 braccia} (‘5-armed candelabrum’) the syntactic pattern \texttt{a N bracci/a} to build multiple variants, as \textit{a 13 bracci}. %
Therefore, models tend to identify productive syntactic patterns and extend them compositionally, rather than drawing on actual distributional evidence or domain knowledge. In essence, \textbf{imitation-based} errors are structural extrapolations that mirror known exemplars too closely, prioritizing form over grounded meaning. 

Additionally, the generated expressions are grammatically well-formed but semantically incoherent, implausible, or internally contradictory. For example, \textit{geranio a foglie di quercia} (‘geranium with oak leaves’) or \textit{a foglie di rosmarino} (‘with rosemary leaves’) attribute biologically implausible features. Similarly, \textit{maglia a punto croce} (‘knitwear in cross-stitch’) is semantically incoherent, because \textit{punto croce} is a specific embroidery technique used to decorate fabrics---not for constructing knitwear. In these cases, LLMs apply compositional plausibility without conceptual coherence: models generate a surface-acceptable phrase that violates domain-specific knowledge or real-world constraints, thereby rendering the expression \textbf{nonsensical}.  
Finally, some generated outputs are not attested exemplars but rather \textbf{novel, \textit{ad hoc} instances} \citep{barsalou1983ad}. For example, the model generates instances of \textit{cassettiera} (‘dresser’) based on spatial context (e.g., \textit{c. da corridoio} ‘hallway dresser’, \textit{c. da esterno} ‘outdoor dresser’) or intended contents (e.g., \textit{c. per giocattoli} ‘for toys’, \textit{per oggetti di cancelleria} ‘for stationery items’). While such expressions might be interpretable and even plausible, they are not attested in usage and do not correspond to established members of the category, i.e., they do not qualify as exemplars stored in long-term memory.

Additional examples of these generative patterns are provided in Tables~\ref{tab:llama_exemplars} and~\ref{tab:halls} (cf. Appendix~\ref{app:hallucinations}). Overall, these examples illustrate how hallucinations often arise from systematic, though flawed, generalization strategies, revealing \textbf{a gap between surface-level fluency and semantic grounding}.

\subsection{Analysis 2: Humans and LLMs Disagree on the Most Available Exemplars}

In the second analysis, we compare the valid exemplars generated by the LLMs with human-generated exemplars. Specifically, we sort both human and LLMs exemplars according to their \textit{availability score}, which reflects the ease with which a word can be produced as a category member (\S \ref{sec:human-dataset-method}). 
Table \ref{tab:topn_matches} reports the results of the intersection between the top-$n$ ($n=\{1,3,5\}$) most available human-generated and machine-generated exemplars, with overlap computed regardless of the production order. The best results are observed for top-5 matches, with \texttt{Nemo-12B} reaching an overlap of 24\% of the generated exemplars. 
The number of matches varies across categories (cf. Appendix \ref{app:study1}). The most significant overlap is observed within the categories of FOODS (\texttt{Nemo-12B}: 37\%, overall: 29\%) and ANIMALS (\texttt{Nemo-12B}: 36\%, overall: 29\%). In contrast, the lowest overlap emerges within the categories BODY PARTS and FURNISHING (\texttt{Nemo-12B}: 16\%, overall: 12\%).

These lower scores may arise for two reasons. First, the model generates valid exemplars, sometimes even matching those produced by humans, but not the most available ones. 
For example,  the top-5 human-generated exemplars of \textit{cane} `dog' (\textit{labrador}, \textit{pastore tedesco} ` German shepherd', \textit{bassotto} `dachshund ', \textit{chihuahua}, \textit{golden retriever}) only partially overlap with those generated by  \texttt{nemo-12B}  (\textit{\underline{pastore tedesco}}, \textit{\underline{golden retriever}}, \textit{beagle}, \textit{\underline{labrador}}, \textit{husky siberiano} `siberian husky').
Besides, \textit{bulldog} is in the top-5 most available exemplars in five models, despite having a lower corpus frequency than other words (e.g., \textit{chihuahua},  \textit{dalmatian}). 
The variation among models suggests that there are \textbf{no specific criteria} (e.g., frequency) \textbf{that determine the generation of one exemplar over another, implying a category organization that is essentially flat}. 

Secondly, some models produce incorrect exemplars: in some cases, meronyms are generated (i.e., \textit{polpaccio} `calf' as a type of \textit{gamba} `leg'), in others,  the basic-level category is misinterpreted due to polysemy (i.e., the word  \textit{braccio} `arm' refers both to a human body part and to an extension of something), resulting in nonsensical outputs. %
Incorrect exemplar generation is especially evident in vLMs. For example, \texttt{idefics2-8B} not only relies on compositional operations but also lists other types of trees (e.g., \textit{acacia, eucalyptus, maple} as exemplars of \textit{abete} `fir'), failing to generate subordinate exemplars and generating basic-level exemplars instead. 

\begin{table}[t!]
    \centering
    \small
    
\begin{tabularx}{\linewidth}{Xccc}
    \toprule
    Model & Top-1 & Top-3 & Top-5 \\
     
    \midrule
    \texttt{llama-3.2-3B}    & 0.09 & 0.13 & 0.14 \\
    \texttt{llama-3.1-8B}    & 0.14 & 0.18 & 0.20 \\
    \texttt{llama-3.1-70B}   & 0.18 & 0.20 & 0.21 \\
    \texttt{mistral-7B}      & 0.13 & 0.12 & 0.13 \\
    \texttt{nemo-12B}        & \textbf{0.25} & \textbf{0.24} & \textbf{0.24} \\
    \texttt{mixtral-8x7B}    & 0.18 & 0.19 & 0.19 \\
    \midrule
    \texttt{llava-7B}        & 0.12 & 0.13 & 0.15 \\
    \texttt{idefics2-8B}     & 0.08 & 0.10 & 0.10 \\
    \bottomrule
    \end{tabularx}

    \caption{Matches among the top-$n$ human and machine-generated most available exemplars.}
    \label{tab:topn_matches}
\end{table}

\section{Are LLMs Sensitive to Human Category Structure?}
\label{sec:tasks}
The comparative analyses of human and LLMs-generated exemplars revealed no significant overlap between these two sets.
However, despite some noisy \textit{ad hoc} exemplars, models also produce valid exemplars that humans did not recall. 
We use human data to build two additional classification tasks:

\begin{enumerate}[label=\Alph*.]
    \item \textbf{Category Induction}:  Given the 10 most available human-generated exemplars, select their basic/superordinate category; 
    \vspace{-.5em}
    \item \textbf{Typicality Detection}: Given the most and least available human-generated exemplars, identify the typical (i.e., most available) member of the basic category.
\end{enumerate}
\vspace{-.5em}
\noindent These tasks are designed to evaluate the model's consistency in representing categories and their exemplars using close-ended formats. Rather than generating exemplars, the model selects correct answers based on its perplexity score, making evaluation easier and more reliable.

\subsection{\textsc{Subtask A}: Category Induction}
\label{sec:super-preds}
Previous studies revealed that basic-level members of a category can elicit the activation of their corresponding superordinate categories in the mental lexicon \cite{barsalou1982context, ross1999food}. 
While tasks in \S \ref{sec:llms-generation} were focused on exemplar generation, here we explore to what extent LLMs are able to identify the category to which an exemplar belongs to. Specifically, we investigate whether subordinate-level members of a given category can activate their \textit{(i)} basic and \textit{(ii)} superordinate category in LLMs.
This allows us to compare recall performances at different levels of taxonomy, from the (more \textit{specific}) basic and (more \textit{general}) superordinate categories, and to better investigate the organization of conceptual categories in the learned latent space of LLMs.

\renewcommand*{\minval}{0.0}%
\renewcommand*{\maxval}{1.0}%

\begin{table}[t!]
\centering
{\small
\begin{tabularx}{\linewidth}{Xcc}
\toprule
Model & Basic-level & Superordinate \\
\midrule
\texttt{llama-3.2-3B}            & 0.84 & 0.52  \\
\texttt{llama-3.1-8B}            & 0.96 & 0.63  \\
\texttt{llama-3.1-70B}           & 0.95 & \textbf{0.64}  \\
\texttt{mistral-7B}              & 0.89 & 0.59  \\
\texttt{nemo-12B}                & 0.95 & 0.46  \\
\texttt{mixtral-8x7B}            & \textbf{0.98} & 0.57  \\
\midrule
\texttt{llava-7B}   & 0.93 & 0.59  \\
\texttt{idefics2-8B}             & 0.94 & 0.38  \\
\bottomrule
\end{tabularx}

}
\caption{\textsc{Subtask A}--Accuracy for basic-level and superordinate category prediction at the aggregated level.
\label{res:subtask-A-aggregate}
}
\end{table}

\paragraph{Setup.} The task is structured as a classification task. Given an input sentence containing a sequence of subordinate-level exemplars, the model has to select the correct category that has produced the listed exemplars.
The category can be: (\textit{i}) one of the 187 basic-level categories (e.g., \textit{abete} `fir', \textit{aereo} 'plane'), or (\textit{ii}) one of the 12 superordinate categories (e.g., \textit{pianta} `plant', \textit{veicolo} `vehicle').
We select up to 10 most available human-generated exemplars for each basic-level concept. Each list is converted into a prompt in the form: ``\texttt{$e_1$, $e_2$,...,$e_{10}$ are types of \{category\}}'', where $e_n$ denotes the $n$-th selected human-produced exemplar and \texttt{category} is a category name, either at basic-level or superordinate one. %
We then compute the model's perplexity for each pair and select the category associated with the sentence that has the lowest perplexity score.

\paragraph{Results.} 
Overall, models obtain higher results when predicting the basic-level concept (e.g., \textit{abete} `fir') rather than the more abstract superordinate category (e.g., \textit{pianta} `plant'; cf. Table \ref{res:subtask-A-aggregate}). This result is surprising, considering that the number of superordinate categories is smaller (12 vs 187 concept terms). A possible explanation is that models have seen the occurrence <exemplar, basic-level concept> more frequently than the pair <exemplar, superordinate-level concept>. In addition, most of the time, the exemplar itself can contain the concept sub-string, e.g., \textit{abete di Natale} (`Christmas tree') vs \textit{?pianta di Natale} (`Christmas plant'). Interestingly, LLM performance varies across semantic domains: models score nearly perfectly on ANIMALS, KITCHEN, and VEHICLES, but perform poorly on FURNISHING, HOBBIES, and STATIONERY (cf. Appendix \ref{app:subtaskA}). As expected,\textbf{ LLMs more effectively acquire taxonomic relations for categories shaped by encyclopedic knowledge} (factual information typically learned through education or texts, e.g., ``a lion is a mammal") than those grounded in commonsense knowledge (e.g. ``domino is a game"). %

\subsection{\textsc{Subtask B}: Typicality Prediction}
\label{sec:typicality}

\renewcommand*{\minval}{0.40}%
\renewcommand*{\maxval}{0.75}%

\begin{table}[t!]
\centering
{\small
\begin{tabularx}{\linewidth}{Xccc}
\toprule
Model & Low & Medium & High \\
\midrule
\texttt{llama-3.2-3B}    & \gradient{0.65} & \gradient{0.62} & \gradient{0.42} \\
\texttt{llama-3.1-8B}    & \gradient{0.58} & \gradient{0.60} & \gradient{0.42} \\
\texttt{llama-3.1-70B}   & \gradient{0.73} & \gradient{0.68} & \gradient{0.61} \\        %
\texttt{mistral-7B}      & \gradient{0.50} & \gradient{0.57} & \gradient{0.47} \\
\texttt{nemo-12B}        & \gradient{0.53} & \gradient{0.69} & \gradient{0.52} \\
\texttt{mixtral-8x7B}    & \gradient{0.72} & \gradient{0.55} & \gradient{0.57} \\
\midrule
\texttt{llava-7B}        & \gradient{0.48} & \gradient{0.62} & \gradient{0.48} \\
\texttt{idefics2-8B}     & \gradient{0.53} & \gradient{0.58} & \gradient{0.45} \\
\bottomrule
\end{tabularx}
}
\caption{\textsc{Subtask B} -- Typicality Accuracy for basic-level categories for the three coverage groupings.}
\label{res:subtask-B-aggregate}

\end{table}

One key aspect of category structure that has been extensively studied with LLMs is \textit{typicality} (\S \ref{subsec:categorystructure-llms}): Some members of a category are considered more representative than others (e.g., \textit{robin} vs. \textit{penguin} as types of birds). 
Previous studies have found only a moderate correlation between human judgments and LLMs. In addition, their focus was basic-level exemplars of superordinate categories. 
In this subtask, we investigate whether, despite their misalignment with humans in generating the most available exemplars (\S\ref{sec:llms-generation}), LLMs can still recognize that the most available item (e.g., \textit{bicchiere di vetro}, `glass tumbler') is more typical than the less available one (e.g., \textit{bicchiere da shot} `shot glass') for a given category (e.g., \textit{bicchiere} `glass'). %

\paragraph{Setup.}
We group the 187 basic-level categories by the number of exemplars produced by humans into three groups: (\textit{i}) \textbf{low} (up to 5 exemplars), (\textit{ii}) \textbf{medium} (6–10 exemplars), and (\textit{iii}) \textbf{high productivity} (more than 10 exemplars). 
This grouping allows us to test if the internal dimension of the category impacts typicality detection results. %
For each basic-level concept, we then select the most available and the least available human-generated exemplars and evaluate the models' perplexity on the two sentences: ``\texttt{\{1st exemplar\} is a type of \{concept\}} vs. \texttt{\{last exemplar\} is a type of \{concept\}}.''
Similarly to \S\ref{sec:super-preds}, a pair is considered a positive prediction if the perplexity for the first sentence is lower than that assigned to the second one.

\renewcommand*{\minval}{0.30}%
\renewcommand*{\maxval}{0.75}%

\begin{table}[t!]
\centering
{\small
\begin{tabularx}{\linewidth}{Xccc}
\toprule
Model & Low $|\Delta|$ & Medium $|\Delta|$ & High $|\Delta|$ \\
\midrule
\texttt{llama-3.2-3B}    & \gradient{0.47} & \gradient{0.58} & \gradient{0.61} \\
\texttt{llama-3.1-8B}    & \gradient{0.45} & \gradient{0.50} & \gradient{0.57} \\
\texttt{llama-3.1-70B}   & \gradient{0.59} & \gradient{0.61} & \gradient{0.70} \\
\texttt{mistral-7B}      & \gradient{0.41} & \gradient{0.49} & \gradient{0.53} \\
\texttt{nemo-12B}        & \gradient{0.43} & \gradient{0.49} & \gradient{0.69} \\
\texttt{mixtral-8x7B}    & \gradient{0.49} & \gradient{0.55} & \gradient{0.69} \\
\midrule
\texttt{llava-7B}        & \gradient{0.42} & \gradient{0.47} & \gradient{0.61} \\
\texttt{idefics2-8B}     & \gradient{0.46} & \gradient{0.47} & \gradient{0.49} \\
\bottomrule
\end{tabularx}
}
\caption{\textsc{Subtask B} -- Typicality Accuracy for basic-level categories, grouped by the absolute difference in exemplars availability.}
\label{res:subtask-B-aggregate-availability}

\end{table}

\paragraph{Results.} 
Overall, \texttt{LLaMA-3.1-70B} performs best across the three groupings, reaching 73\% accuracy in the low-productivity setting (cf. Table \ref{res:subtask-B-aggregate}), a good score compared to past studies. However, accuracy varies across groupings: \textbf{as the number of human exemplars for a category increases, LLMs are less likely to detect the typical item.}
This suggests that when humans provide fewer exemplars, the first one is cognitively dominant compared to the other ones, a distinction reflected in the model’s perplexity scores. However, in richer categories, the cognitive distinctive attributes among exemplars diminish, thus resulting in LLMs' lower accuracies (cf. Appendix \ref{app:subtaskB}).

\paragraph{Effect of Availability Differences.}
Additionally, we assess accuracy across groups defined by the absolute difference in availability ($|\Delta|$) between the most and least available exemplars.  We categorize these differences into three levels: \textbf{low $\Delta$} for differences less than $0.2$, \textbf{high} $\Delta$ for differences greater than $0.4$, and \textbf{medium} $\Delta$ for all other cases. This grouping results in a balanced distribution of pairs (57, 75, and 55, respectively).
Looking at the average results in Table \ref{res:subtask-B-aggregate-availability}, we observe that \textbf{pairs with a higher typicality delta are easier to predict}, yielding higher accuracy scores. For example, the best performing model \texttt{LLaMA-3.1-70B} achieves almost a 20\% increase when moving from the low to the high $\Delta$ setting (and a $\sim$30\% on average across all the models). 
This additional analysis reveals that LLMs are sensitive to the internal structure of human basic-level categories: \textbf{the smaller the variability in human availability}, %
\textbf{the more difficult it becomes for the model to identify the most typical items}.

\section{General Discussion and Conclusions}
\label{sec:discussion}

This study explored basic-level category organization in humans, who integrate linguistic and sensory information, and LLMs, which rely solely on linguistic data. In a generation task, Italian speakers and various LLMs and vLMs produced lists of exemplars for 187 basic-level concrete categories. We hypothesized that the most frequent exemplars generated by models would align with those of humans, as subordinate concepts reflect specialized knowledge and are constrained by language. 

  Findings in \S \ref{sec:llms-generation}  reveal a low alignment between model and human performance. However, comparative analyses show that some models (particularly \texttt{LLaMA-3.1-70B}) can still generate meaningful exemplars comparable to those produced by humans across many semantic domains. Interestingly, these models produce more exemplars than humans for technical and specialized categories that require access to encyclopedic knowledge (i.e., PLANTS): e.g.,  \texttt{LLaMA-3.1-70B} generates 26 real exemplars for \textit{orchidea} `orchid', while humans generated only 5. This ability points to a possible use of LLMs in automatically generating exemplars for large sets of concepts (i.e., for automatic ontology population), in line with similar findings for semantic feature production norms \citep{Hansen2022}. However, our results also call for some caution.
  
First, the models often generate hallucinations and incorrect exemplars, especially for categories where extralinguistic information plays a more critical role than linguistic data. This is especially evident in the BODY PARTS category, where conceptual confusion (\textit{piede di porco} ‘crowbar’) or\textit{ ad hoc} instances (\textit{testa di cavallo} ‘horse head’) are common.  %
While frequency analysis can help reduce hallucinations, human annotation is needed to verify accuracy, at least at this taxonomic conceptual level.
Secondly, LLMs do not show the same categorical organization of humans. %
The generated exemplars vary significantly across models, with alignment to human responses below 25\% (\S \ref{sec:llms-generation}). 

Additional subtasks in \S \ref{sec:tasks} illustrate that models struggle to build a hierarchical conceptual organization like humans, limiting their ability to reason along the taxonomic axis (\S\ref{sec:super-preds}). While they perform well in basic-level category induction, they underperform in the superordinate category setting. 
Moreover,  LLMs often fail to identify the most typical exemplar when a category includes multiple similarly available items (\S\ref{sec:typicality}) but perform better when one exemplar clearly dominates in availability. These results suggest that (proto)typicality effects are harder to detect within basic-level categories, likely due to their relatively flat internal structure and the high number of shared attributes among subordinate exemplars. 
Finally, we found that vLMs still perform poorly in the exemplar generation task, in line with previous research \citep{vemuri2024well}, showing that text-based models align more closely with human typicality judgments.

Our study has several methodological implications worth mentioning. We provided a dataset of human-generated exemplars for basic-level concrete categories in Italian, along with statistical measures, extending \citet{Montefinese2012SemanticMA}. Since existing Italian datasets often lack concepts spanning multiple taxonomic levels, this resource will be useful in cognitive psychology and AI research on semantic category structure. This need for comprehended datasets becomes evident when comparing existing resources in other languages, such as English (e.g., \citealp{banks2023category}).  Moreover, our study highlights the potential and limitations of LLMs in capturing human categorical knowledge at the subordinate level, in line with previous literature. 
Future work should explore how LLMs generate exemplars for superordinate categories (e.g., \textit{animals, plants}) and whether they align more with human behaviour at this level. Additionally, comparing results across languages could also reveal cultural influences on concept representation and potential biases in LLMs.

In conclusion, our results show that the organization of subordinate categories varies as a function of semantic domains in both humans and LLMs. Notably, the more extralinguistic or linguistic information is relevant to a given category, the more the performance of LLMs and humans diverges. 
These observations have practical implications for NLP systems, such as educational tools (e.g., vocabulary teaching, interactive learning apps), knowledge base population, and generally, to improve category-aware language generation (i.e., chatbots that better interpret user intent by responding with the appropriate level of specificity).

\label{sec:limitations}
\section*{Limitations}
\begin{enumerate}
    \item \textbf{Cultural Biases}: Model are trained on English and/or multilingual corpora which may not reflect the lexical preferences of Italian speakers.
    \item \textbf{Methodology in 4.2}: In the comparison between LLMs and human-generated exemplars, we used a simple string matching, so \textit{abete di Natale} `Christmas fir' and \textit{abeti di Natale} `Christmas firs' are considered different strings. While this approach could count good strings as mismatches, the human judgments are manually normalized, and models prefer the singular form consistently. In conclusion, we believe that this approximation does not exclude too many possibly good exemplars.
    \item \textbf{Exclude GPT from analyses}: We did not use GPT because we cannot access the perplexity values of the model. While some could argue that GPT last models could achieve better performances for the presented tasks, we prefer open models that can be accessed in their internal representations.
\end{enumerate}

\section*{Ethical Considerations}
\begin{itemize}
    \item We administrated the exemplars generation task described in \S \ref{sec:human-dataset} to a total of 365 participants (48.5\% women; 49.9\% man; 1.6\% non-binary; M age = 26.3; SD age = 3.76; range age 18-35) on Prolific. All participants were Italian native speakers and reported no language or attentional disorders. Participants were compensated with Euro \EUR{1.80} for generating exemplars in a single list, with an average survey duration of 15 minutes.  The data is anonymized to make identification of individuals impossible.

    \item Since the human data were collected in 2023 and never released, all LLMs have not been exposed to these stimuli, allowing us to test the emerging abilities of these models and their semantic knowledge. 
    
    \item This research demonstrates the utility of language models as valuable tools in cognitive science and linguistics. However, it is crucial to acknowledge that these models acquire and produce language through mechanisms that differ significantly from human language processing. Consequently, extrapolating these findings directly to human mind organization can lead to potential risks and unintended consequences.
\end{itemize}

\section*{Acknowledgements}
We would like to thank the anonymous reviewers for their feedback and comments.
AP has been supported by the project ‘‘Word Embeddings: From Cognitive Linguistics to Language Engineering, and Back’’ (WEMB), funded by the Italian Ministry of University and Research (MUR) under the PRIN 2022 funding scheme (CUP B53D23013050006),
the PNRR (Prot. IR0000013) ‘‘SoBigData.it: Strengthening the Italian RI for Social Mining and Big Data Analytics’’. 
GR, CV, and MB have been funded by the European Union  (GRANT AGREEMENT: ERC-2021-STG-101039777). Views and opinions expressed are however those of the author(s) only and do not necessarily reflect those of the European Union or the European Research Council Executive Agency. Neither the European Union nor the granting authority can be held responsible for them. 

\bibliography{anthology, custom}
\bibliographystyle{acl_natbib}
\appendix
\section{\textsc{Study 1}}
\subsection{Metrics}
\label{app:human-metrics}
In this section, we define the metrics described in Section \ref{sec:human-dataset} used to evaluate the exemplars obtained from human participants.

\paragraph{Exemplar Dominance}
\begin{equation}
    \textit{ED}(E) = P(E|C) = \frac{N(E \cap C)}{N(C)}
\end{equation}
\noindent where $N(E \cap C)$ is equal to the number of participants who produced the exemplar $E$ when in response to the concept $C$, and $N(C)$ is the number of participants elicited by $C$.

\paragraph{Mean Rank Order}
\begin{equation}
    \textit{MRO}(E) = \frac{\sum^{N(C)} r_i(E|C)}{N(C)}
\end{equation}

\paragraph{First Occurrence Value}
\begin{equation}
    \textit{FOV}(E, C) = \frac{N_{first}(E)}{N(C)}
\end{equation}

\paragraph{Exemplar Availability}
\begin{equation}
    \textit{EA(E,C)} = \sum_{p=1}^{n}\frac{f_{pi}}{N}\cdot e^{[-2.3\cdot(\frac{p-1}{n-1}])}
\end{equation}
\noindent where $p$ is the rank of the produced exemplar $E$, $n$ is its lowest rank obtained across multiple participants, $f_{pi}$ is the number of participants who produced the exemplar $i$ at the same position $p$, and $N$ is the total number of participant who have seen the category $C$.

\section{\textsc{Study 2}}
\subsection{Models Description}
\label{app:models}
In this section, we provide the details on the pretrained language models listed in \S \ref{sec:llms-generation}. All models are open-source and available via huggingface\footnote{\url{https://huggingface.co/}}.
\subsection{Unimodal Language Models}
\paragraph{LLaMA-3.1} \cite{dubey2024llama3herdmodels} is a collection of pre-trained auto-regressive large language models openly released by Meta AI. In our experiments, we rely on the instruction-based version, which are fine-tuned for dialogue use case with multilingual input. We assess performance of both the small version (8B parameters\footnote{\url{https://huggingface.co/meta-llama/Llama-3.1-8B-Instruct}}) and the larger one (with 70B parameters\footnote{\url{https://huggingface.co/meta-llama/Llama-3.1-70B-Instruct}}). We avoid testing the extra-large version (405B parameters) due to computational constraints. All models are first pre-trained (SFT) on a mix of publicly available online data and further aligned with human preferences via RLHF.

\paragraph{LLaMA-3.2} is the next iteration of llama models. With respect to version 3.1, they differ in models sizes (1B, 3B, 11B, and 90B parameters) and multimodal capabilities. However, at the moment of writing, the multimodal version of llama-3.2 is not accessible in the EU, due to European regulations\footnote{\url{https://huggingface.co/meta-llama/Llama-3.2-11B-Vision-Instruct}}. For this reason, we are not able to provide any insight about the multimodal versions. Concerning the assessed version, we limit ourselves to the small (3B) model.\footnote{\url{https://huggingface.co/meta-llama/Llama-3.2-3B-Instruct}}

\paragraph{Mistral} \cite{jiang2023mistral7b} is a pre-trained auto-regressive large language model released by Mitral AI\footnote{\url{https://huggingface.co/mistralai/Mistral-7B-Instruct-v0.2}}. The model leverages Grouped-Query Attention and Sliding Windows Attention to improve inference time and memory requirements, and to enable handling longer input sequences. 

\paragraph{Mistral-8x7B} is an ensemble mixture of experts model\footnote{\url{https://huggingface.co/mistral/Mistral-7B-Instruct}} of eight 7B parameter models developed by Mistral AI. The individual models are trained with Grouped-Query Attention (GQA) and Sliding Window Attention (SWA) mechanisms, enabling efficient handling of long sequences and improving inference speed. A routing system takes care of distributing the input to the appropriate experts. This mechanism increases the number of parameters of a model while controlling cost and latency, as the model only uses a fraction of the total set of parameters per token. 
For our experiments, we use the standard instruction-tuned version of Mistral-7B, focusing on its capacity for multilingual inputs and dialogue generation. 

\paragraph{NeMo} is a 12B model\footnote{\url{https://huggingface.co/mistralai/Mistral-Nemo-Base-2407}} designed for multilingual applications. It is trained on function calling, has a large context window, and is particularly strong in English, French, German, Spanish, Italian, Portuguese, Chinese, Japanese, Korean, Arabic, and Hindi. Mistral NeMo uses a new tokenizer, Tekken, based on Tiktoken, that was trained on over more than 100 languages, and compresses natural language text and source code more efficiently than the SentencePiece tokenizer used in previous Mistral models.

\subsection{Multimodal Language Models}

\paragraph{LLaVA} \cite{liu2024visual} is a multimodal model that integrates visual understanding with language capabilities by combining a vision encoder (e.g., CLIP’s Vision Transformer) with a large language model (e.g., LLaMA). It is designed for open-ended vision-language tasks, such as image captioning, visual question answering, and reasoning about images.
The model is trained following a two-stage training approach: first, the vision and the language encoders are aligned by training a projection layers that map visual features into the LLM's embedding space. Second, the model undergoes an instruction tuning phase, using curated vision-language datasets to improve coherence and accuracy in responses.

\paragraph{Idefics2} \cite{laurenccon2024matters} is the result of a throughout ablation of the design choices available for vLMs pre-training. To encode visual features in the LLM's embedding space, Idefics2 leverages a SigLIP's vision encoder \cite{zhai2023siglip} followed by a learned Perceiver pooling \cite{jaegle2021perceiver} and an multi-layer perceptron projection. The pooled sequence is then concatenated with the text embeddings to obtain an interleaved sequence of images and texts. The model is trained according to the usual vLMs pipeline, with a first stage focusing on the alignment of the two modality embedders, followed by a second instruction-tuning stage.

\subsection{Perplexity Computation}
Perplexity is computed according the following formula:
\begin{equation}
    \text{PPL}(X) = \exp\left\{ \frac{1}{t} \sum_i^t \log p_{\theta}(x_i \mid x_{<i}) \right\}
\end{equation}

\noindent where $x_i$ is the target expression (i.e., either the basic or superordinate category, in \textsc{Subtask A}, or the subordinate level exemplar, in \textsc{Subtask B})  and $x_{<i}$ is the fixed prompt. In our settings, this is equivalent to the exponentiation of the cross-entropy loss. We compute the perplexity for the target tokens only $(x_i)$, and mask the non-target tokens $(x_{(<i)})$ accordingly. Notice that in our experiments the perplexity is used to compare output of the same model, therefore normalization is not required to compare the binary-accuracy score (i.e., the evaluation metrics for \textsc{Subtask A} and \textsc{B}).

\subsection{Prompting Strategy}
\label{app:llm-gen-prompt}
To obtain a list of exemplars (i.e., basic-level concepts) from a LLMs, we use the following Italian prompt:

\begin{quote}
    \texttt{<s>[INST] Data una parola che denota una concetto, elenca tutta i `tipi di' quel concetto. Elenca solo i nomi delle entità. Per esempio per il concetto `elettrodomestico' elenca: frullatore, aspirapolvere, tostapane, lavatrice. Ora fai lo stesso per il concetto `<CONCEPT>' [/INST] Questa è una lista 'tipi di' che appartengono al concetto `<CONCEPT>`:}
\end{quote}
where \texttt{<CONCEPT>} is replaced with the eliciting concept. For the non-Italian reader, we provide an English translation of previous prompt:

\begin{quote}
    \texttt{<s>[INST] Given a word denoting a concept, list all of the `kinds of' of the given concept. List only words denoting entities. For example, for the concept `electric appliance` list: `mixer', `vacuum cleaner', `toaster', `washing machine'. Now do the same for the concept `<CONCEPT>':}
\end{quote}

\renewcommand*{\minval}{0.00}%
\renewcommand*{\maxval}{0.45}%

\begin{table*}[t!]
\resizebox{\linewidth}{!}{%

\begin{tabular}{lccccccccccccr}
\toprule
 & \MakeUppercase{animals} & \MakeUppercase{body parts} & \MakeUppercase{clothes} & \MakeUppercase{foods} & \MakeUppercase{furnishing} & \MakeUppercase{furniture} & \MakeUppercase{hobbies} & \MakeUppercase{housing} & \MakeUppercase{kitchen} & \MakeUppercase{plants} & \MakeUppercase{stationery} & \MakeUppercase{vehicles} & avg \\
 
\midrule
\texttt{llama-3.2-3B} & \gradient{0.24}& \gradient{0.13}& \gradient{0.09}& \gradient{0.33}& \gradient{0.11}& \gradient{0.13}& \gradient{0.08}& \gradient{0.10}& \gradient{0.12}& \gradient{0.06}& \gradient{0.06}& \gradient{0.21}& 0.14\\
\texttt{llama-3.1-8B} & \gradient{0.32}& \gradient{0.13}& \gradient{0.09}& \gradient{0.36}& \gradient{0.20}& \gradient{0.24}& \gradient{0.18}& \gradient{0.19}& \gradient{0.17}& \gradient{0.15}& \gradient{0.13}& \gradient{0.25}& 0.20\\
\texttt{llama-3.1-70B} & \gradient{0.35}& \gradient{0.12}& \gradient{0.15}& \gradient{0.29}& \gradient{0.19}& \gradient{0.28}& \gradient{0.23}& \gradient{0.19}& \gradient{0.15}& \gradient{0.18}& \gradient{0.18}& \gradient{0.18}& 0.21\\
\texttt{mistral-7B} & \gradient{0.25}& \gradient{0.14}& \gradient{0.07}& \gradient{0.24}& \gradient{0.03}& \gradient{0.09}& \gradient{0.14}& \gradient{0.13}& \gradient{0.13}& \gradient{0.16}& \gradient{0.08}& \gradient{0.13}& 0.13\\
\texttt{nemo-12B} & \gradient{0.36}& \gradient{0.16}& \gradient{0.26}& \gradient{0.37}& \gradient{0.16}& \gradient{0.31}& \gradient{0.26}& \gradient{0.18}& \gradient{0.23}& \gradient{0.24}& \gradient{0.25}& \gradient{0.15}& \textbf{0.24}\\
\texttt{mixtral-8x7B} & \gradient{0.27}& \gradient{0.11}& \gradient{0.18}& \gradient{0.25}& \gradient{0.19}& \gradient{0.20}& \gradient{0.23}& \gradient{0.19}& \gradient{0.18}& \gradient{0.18}& \gradient{0.21}& \gradient{0.14}& 0.19\\
\midrule
\texttt{llava-7B} & \gradient{0.32}& \gradient{0.09}& \gradient{0.11}& \gradient{0.28}& \gradient{0.06}& \gradient{0.11}& \gradient{0.15}& \gradient{0.10}& \gradient{0.10}& \gradient{0.22}& \gradient{0.10}& \gradient{0.15}& 0.15\\
\texttt{idefics2-8B} & \gradient{0.25}& \gradient{0.09}& \gradient{0.06}& \gradient{0.25}& \gradient{0.03}& \gradient{0.04}& \gradient{0.11}& \gradient{0.03}& \gradient{0.03}& \gradient{0.10}& \gradient{0.03}& \gradient{0.21}& 0.10\\
\bottomrule
category avg & \textbf{0.29} & 0.12 & 0.12 & \textbf{0.29} & 0.12 & 0.17 & 0.17 & 0.13 & 0.13 & 0.16 & 0.13 & 0.17 & 0.17\\

\bottomrule
\end{tabular}

}

\caption{Percentage of matches among \textbf{top five} most available exemplars.}\label{tab:top5matches_categories}

\end{table*}

\subsection{Model-specific sampling parameters}
Regarding hyperparameters, we set \texttt{top-p} to 0.75 to limit the long tail of low-probability tokens that may be sampled, while \texttt{frequency} and \texttt{repetition penalty} are set to 0.

\subsection{Top-5 Matches}\label{app:study1}
Table \ref{tab:top5matches_categories} shows the percentage of matches among the top-5 human-
produced and LLMs-generated exemplars, reporting individual accuracy for each of the 12 superordinate categories.

\subsection{Generated Exemplars and Hallucinations}\label{app:hallucinations}
In Table \ref{tab:llama_exemplars} we report the exemplars generated by the LLaMA-3.1-70B, the best-performing model, for the 12 superordinate categories. For each of the 12 superordinate categories, we select the basic-level concept for which humans
have generated the greatest amount of exemplars. In Table \ref{tab:halls}, we report the exemplars generated for the 12 basic-level concepts that produced the greatest amount of unattested occurrences according to the Italian Corpus ItTenTen.

In our study, we automatically identify low-frequency occurring terms via the Italian corpus ItTenTen. By analyzing exemplars with an absolute frequency equal to zero we can gain a deeper insight regarding hallucination generation in the exemplars generation task. 
We divide unattested exemplars into false negatives (e.g., exemplars for which we retrieved a zero frequency due to misspellings or morphosyntactical issues) and hallucinations.
Through qualitative analysis, we observe several recurring patterns and categorize most of the hallucinations into the following groupings: \textit{ad-hoc instances}, \textit{nonsensical}, \textit{foreign-language based}, \textit{conceptual confusion}, and \textit{imitation-based}.

\paragraph{\textit{Ad-hoc} Instances:} These instances reflect the model’s ability to creatively compose category-consistent yet ungrounded expressions, relying on syntactic and semantic cues rather than empirical knowledge. As such, ad hoc constructions are generated ``on the fly'' to fit perceived communicative goals, but lack the frequency-based support or conventionalization required to qualify as exemplars stored in long-term memory. Some examples are: MAGLIA `a punto catenella' (chain stitched KNITWEAR), `a punto scritto a rombi' (diamond shape stitched KNITWEAR), GALLO `della giungla verde' (COCK of the green jungle), `della giungla rosso' (red COCK of the jungle), CASSETTIERA `per giocattoli' (toy DRAWER), or CASSETTIRA `per attrezzi' (tool DRAWER),  `da corridoio' (hallway DRAWER).

\paragraph{Nonsensical:} Expressions that are grammatically well-formed but semantically incoherent, implausible, or internally contradictory, often resulting from incongruous or incompatible feature combinations. Some examples are: GERANIO `a foglie di quercia' (GERANIUM with oak leaves), `a foglie di rosmarino' (GERANIUM with rosemary leaves). CRUCIVERBA `a parole sovrapposte' (CROSSWORD with overlapping words), `a parole crociate' (CROSSWORD with word crossed). TRATTORE `a cingoli in acciaio' (TRACTOR with steel tank track). GALLO `cedrone giapponese' (Japanese capercaillie COCK).

\paragraph{Foreign-Language Based:} Refers to expressions that denote a real-world referent conceptualized in a foreign language with respect to Italian. For example, GALLO `di Crèvecœur' (Crèvecœur CHICKEN) has no attested translation in Italian.

\begin{table*}
    \definecolor{lightgreen}{RGB}{200,255,200}
\definecolor{lightred}{RGB}{255,200,200}
\definecolor{lightyellow}{RGB}{255,255,200}
\definecolor{lightblue}{RGB}{200,200,255}

\definecolor{red}{RGB}{255,0,0}               %
\definecolor{blue}{RGB}{0,0,255}              %

\newcommand{\mis}{\cellcolor{lightblue}}
\newcommand{\low}{\cellcolor{lightyellow}}
\newcommand{\hal}{\cellcolor{lightred}}

\resizebox{\textwidth}{!}{
\begin{tabular}{lllllll}

\toprule
 & \MakeUppercase{animals} & \MakeUppercase{body parts} & \MakeUppercase{clothes} & \MakeUppercase{foods} & \MakeUppercase{furnishing} & \MakeUppercase{furniture} \\
&               \textbf{Cane} (dog)  &                  \textbf{Capelli} (hair) &                          \textbf{Scarpa} (shoe) &             \textbf{Pasta} (pasta) &                     \textbf{Vaso} (vase) &                          \textbf{Sedia} (chair) \\
\midrule

1 & pastore tedesco & riccio & \mis stivale & spaghetti & \mis di fiori & poltrona \\
2 & \mis segugio & \mis ricci & \mis sandalo & fettuccine & \mis di rame & a dondolo \\
3 & rottweiler & \mis afro & ciabatta & penne & \mis di cristallo & a rotelle \\
4 & alano & \mis ondulato & \mis da ballo & farfalle & di ceramica & sgabello \\
5 & dobermann & \mis crespo & \mis anfibio & tortellini & \mis di porcellana & pieghevole \\
6 & levriero & \mis riccio afro & \mis stivaletto & rigatoni & di terracotta & \mis sdraio \\
7 & corso & \mis liscio lucido & da trekking & cannelloni & di vetro & da giardino \\
8 & pinscher & liscio & da ginnastica & ravioli & \mis di metallo & da ufficio \\
9 & boxer & \mis ondulati & da calcio & gnocchi & \mis urna & a sdraio \\
10 & beagle & \mis crespi & da tennis & maccheroni & \mis di legno & \mis da bar \\
11 & \mis poodle & \mis ricciolino & \mis zoccolo & lasagne & \mis di marmo & \mis da ristorante \\
12 & \mis pug & \mis liscio opaco & \mis da sci & \mis vermicelli & di plastica & \mis da spiaggia \\
13 & dalmata & mosso & mocassino & tagliatelle & \mis di argento & reclinabile \\
14 & bulldog & \mis mossi & \mis da ciclismo & fusilli & \mis di oro & \mis per bambini \\
15 & \mis lupo & \mis lisci & \low da danza classica & linguine & \low di ottone & \low a sacco \\
16 & \mis shih tzu & \hal crespo lucido & \mis da basket & ditalini & \mis di pietra & \mis a schienale alto \\
17 & pitbull & \low ondulato lucido & \mis da neve & pappardelle & \mis di argilla & \low a schienale basso \\
18 & \mis basset hound & \low mosso lucido & \mis da danza & orecchiette & \low di notte & \mis pouf \\
19 & chihuahua & \hal riccio afro lucido & \mis da calcetto & conchiglie & \mis di bronzo & \mis panchina \\
20 & \mis collie & \low riccio lucido & \low da equitazione & \mis lasagna & greco & \mis a braccioli \\

\bottomrule
\end{tabular}}

    \vspace{0.5cm}

\resizebox{\textwidth}{!}{
\begin{tabular}{lllllll}

\toprule
 & \MakeUppercase{hobbies} & \MakeUppercase{housing} & \MakeUppercase{kitchen} & \MakeUppercase{plants} & \MakeUppercase{stationery} & \MakeUppercase{vehicles} \\
&               \textbf{Libro} (book) &                  \textbf{Camera} (room) &                          \textbf{Pentola} (pan) &              \textbf{Margherita} (daisy) &                     \textbf{Foglio} (sheet) &                          \textbf{Automobile} (car) \\
\midrule

1 & romanzo & da letto & a pressione & comune & di carta & berlina \\
2 & saggio & oscura & casseruola & \mis dei prati & \mis di via & \mis autocarro \\
3 & dizionario & \mis di equilibrio & \mis di ghisa & \mis di savoia & di alluminio & \mis autobus \\
4 & \mis atlante & d'albergo & padella & pizza & \mis di rame & \mis camion \\
5 & enciclopedia & \mis iperbarica & \mis marmitta & \mis di lorena & \mis di calcolo & \mis minivan \\
6 & \mis agenda & \mis di sicurezza & wok & \low di angoulême & \mis di stile & monovolume \\
7 & manuale & \mis di combustione & di acciaio & \mis di borgogna & \mis elettronico & \mis spider \\
8 & \mis almanacco & a gas & \mis a vapore & \mis di fiandra & \hal di ruta & suv \\
9 & biografia & \mis di commercio & di rame & \mis di scozia & di plastica & \mis pick-up \\
10 & fumetto & \mis di decompressione & di terracotta & \mis di provenza & \mis di piombo & \mis cabriolet \\
11 & \mis trattato & \mis di scoppio & \mis di ceramica & \mis di parma & \mis di registro & station wagon \\
12 & diario & \mis di refrigerazione & \mis di alluminio & \mis di valois & \hal di viti & fuoristrada \\
13 & \mis catalogo & \mis di consiglio & antiaderente & \mis tudor & \mis di stagno & \mis furgone \\
14 & \mis novella & \mis di compensazione & \mis di pietra ollare & \hal d'ungheria & \mis di ottone & citycar \\
15 & \mis autobiografia & \mis di controllo & \mis coccotte & \hal a fiori doppi & \hal di rame berillifero & \mis furgoncino \\
16 & \mis compendio & \mis di manovra & \mis elettrica & \hal d'austria & \hal di rame fosforoso & coupé \\
17 & \mis raccolta & \mis di decantazione & \mis in rame & \hal a fiori giganti & \hal di rame arsenicale & \mis hatchback \\
18 & per bambini & \mis di carico & \mis in pietra ollare & \hal a fiori piccoli & \mis di stagnola & \mis pulmino \\
19 & \mis racconto & \mis di regia & \mis paiolo & \hal a fiori colorati & \hal di alluminio stagnato & \mis autovettura \\
20 & \mis monografia & \mis mortuaria & \mis calderone & \hal a fiori bianchi & \mis di lavoro & \mis roadster \\

\bottomrule
\end{tabular}}

    \caption{Up to 20 exemplars generated by LLaMA-3.1-70B (the best-performing model in terms of valid exemplars generated), sorted by availability score. For each of the 12 superordinate categories (in \MakeUppercase{uppercase}), we select the basic-level category (in \textbf{bold}) for which humans have generated the greatest amount of exemplars. 
    Cells with a \colorbox{lightblue}{\textbf{light-blue background}} indicate exemplars not produced by the human study group but still considered valid, with more than 15 occurrences in the ItTenTen corpus. Exemplars with lower frequency are denoted by a \colorbox{lightyellow}{\textbf{light-yellow background}}. A \colorbox{lightred}{\textbf{light-red background}} indicates unattested exemplars, which are regarded as hallucinations.}
    \label{tab:llama_exemplars}
\end{table*}

\begin{table*}[t!]
    \definecolor{lightgreen}{RGB}{200,255,200}
\definecolor{lightred}{RGB}{255,200,200}
\definecolor{lightyellow}{RGB}{255,255,200}
\definecolor{lightblue}{RGB}{200,200,255}

\definecolor{red}{RGB}{255,0,0}               %
\definecolor{blue}{RGB}{0,0,255}              %

\newcommand{\mis}{\cellcolor{lightblue}}
\newcommand{\low}{\cellcolor{lightyellow}}
\newcommand{\hal}{\cellcolor{lightred}}

\resizebox{\textwidth}{!}{
\begin{tabular}{lllllll}

\toprule
 & \MakeUppercase{animals} & \MakeUppercase{body parts} & \MakeUppercase{clothes} & \MakeUppercase{foods} & \MakeUppercase{furnishing} & \MakeUppercase{furniture} \\
&               \textbf{Gallo} (cock)  &                  \textbf{Spalla} (shoulder) &                          \textbf{Maglia} (sweater) &             \textbf{Latte} (milk) &                     \textbf{Candelabro} (candelabra) &                          \textbf{Cassettiera} (dresser) \\
\midrule

1 & cedrone & \low a sbalzo & \mis a coste & di cocco & da tavolo & da ufficio \\
2 & \mis bankiva & \hal a volant & \hal a punto croce & di soia & \low a sospensione & \low da cucina \\
3 & \mis silvestre & \low a bretella & a righe & di capra & \low da terra & \low da bagno \\
4 & nero & \hal a bretelle & \low a losanghe & di mucca & \low da parete & \low da notte \\
5 & \hal di banca & \low a botte & \mis rasata & di pecora & \low a 5 bracci & \low da camera da letto \\
6 & \hal di wallich & \low a spigolo & \low a uncinetto & di bufala & \mis a 3 braccia & \hal per giocattoli \\
7 & \mis da combattimento & \hal a pizzo & \mis a punto & \mis di avena & \low a stelo & \low da ingresso \\
8 & \low di sonnerat & \hal a cuscino & \low a tubolare & \low di arachidi & \low a 5 braccia & \hal da comodino \\
9 & \mis della giungla & \hal all'americana & \mis a rombi & di mandorla & \mis a 7 braccia & \hal per attrezzi \\
10 & \hal cedrone giapponese & \low a kimono & \low a cavi & di riso & \mis a 9 braccia & \mis da scrivania \\
11 & \hal di faverolles & \low a punta & \low a fantasia & \mis di cammello & \mis a 7 bracci & \hal per oggetti di cancelleria \\
12 & \hal di houdan & \low a sbuffo & \hal a doppia punta & \mis di nocciole & \mis a 9 bracci & \hal da corridoio \\
13 & \hal della malesia & \hal a frangia & \hal a punto catenella & \mis di anacardi & \hal a 11 bracci & \hal da esterno \\
14 & \hal della giungla grigio & \hal a pizzo di sanok & \hal a punto lino & \mis di quinoa & \hal a 13 bracci & \low da soggiorno \\
15 & \hal di crèvecoeur & \hal a pizzo di lefkara & \hal a punto scritto & \mis di mandorle & \hal da mensola &  \\
16 & \hal della giungla verde & \hal a latticciolo & \hal a punto raso & \low di orzo & \hal da camino &  \\
17 & \hal della giungla rosso & \hal a piquet & \mis a punto legaccio & \low di semi di lino & \low da altare &  \\
18 & \hal di jungla & \hal a pizzo di burano & \hal a punto scritto a rombi & \mis di grano & \hal tripode &  \\
19 & \hal di borneo & \hal a pizzo di gorizia & \hal a punto reale & \low di semi di sesamo & \low da chiesa &  \\
20 & \hal di delacour & \hal a pizzo ricamato & \low a punto rovescio & \low di semi di girasole & \low a più bracci &  \\

\bottomrule
\end{tabular}}

    \vspace{0.5cm}
    
    \resizebox{\textwidth}{!}{
\begin{tabular}{lllllll}

\toprule
 & \MakeUppercase{hobbies} & \MakeUppercase{housing} & \MakeUppercase{kitchen} & \MakeUppercase{plants} & \MakeUppercase{stationery} & \MakeUppercase{vehicles} \\
&               \textbf{Cruciverba} (crossword) &                  \textbf{Terrazzo} (terrace) &                          \textbf{Mestolo} (ladle) &              \textbf{Gernaio} (geranium) &                     \textbf{Colla} (glue) &                          \textbf{Trattore} (tractor) \\
\midrule

1 & \low classico & \mis alla veneziana & \mis forato & \hal a foglie di quercia & a caldo & agricolo \\
2 & \mis enigmistica & \mis a sbalzo & \mis da minestra & \low geranium maculatum & vinilica & cingolato \\
3 & \hal a schema variabile & \hal alla romana & \low lungo & \hal a fioritura continua & \mis a freddo & \mis a ruote \\
4 & a schema fisso & \mis pensile & \low a buco & \hal a foglie di vite & \low a base di lattice & \mis stradale \\
5 & a schema libero & \hal alla genovese & \low da zuppa & \hal a foglie di betulla & \low a base di gomma & \low a cingoli in gomma \\
6 & \mis per bambini & \hal alla milanese & \mis da cucina & \hal a foglie di rosmarino & \low a base di resina & \mis articolato \\
7 & \low tematico & \hal alla bergamasca & \hal da gelato & \hal a foglie di alloro & \hal a base di silicio & \mis a cingoli \\
8 & \hal per adulti & \mis a livello & \low per mescolare & \hal a foglie di felce & \mis epossidica & \low telescopico \\
9 & \hal a schema personalizzato & \hal alla pavese & \low per servire & \hal a foglie di platano & \mis per legno & \mis a quattro ruote motrici \\
10 & \hal a schema geometrico & panoramico & \low per gelato & \hal a fioritura estiva & \mis per carta & \low a ruote motrici \\
11 & \hal a schema logico & coperto & \hal da legno & \hal a fioritura primaverile & \mis per plastica & \hal a cingoli motrici \\
12 & \hal a schema numerico & \mis fiorito & \low da salsa & \hal d'appartamento & \mis acrilica & \mis agricolo cingolato \\
13 & \hal a parole sovrapposte & \mis scoperto & \hal da metallo & \hal d'altura & \hal a base di silano & \hal a cingoli motrici 4x2 \\
14 & \hal a parole nascoste & \mis giardino & \hal da risotto & \hal cespuglioso & \hal a base di silice & \hal a cingoli motrici 4x4 \\
15 & \hal a parole crociate & \mis solarium & \hal da silicone & \low bicolore & \mis per tessuti & \hal a ruote motrici 4x2 \\
16 & \hal a definizioni consecutive & \mis adiacente & \hal per impastare & \hal aquilegifolium & \hal a base di silicato & \hal a ruote motrici 4x4 \\
17 & \hal a definizioni incrociate & \hal alla fiorentina & \low per dosare & \low annuale & \low per metalli & \hal a ruote anteriori sterzanti \\
18 & \hal a tema libero &  & \hal per condire & \hal alpino & \mis per vetro & \hal a ruote posteriori sterzanti \\
19 & \hal a figure &  & \low cucchiaio & \hal a foglia rossa & \low a base di solvente & \low a due ruote motrici \\
20 & \low con immagini &  & \hal a nido d'ape & \low geranium phaeum & \hal a base d'acqua & \hal a cingoli in acciaio \\

\bottomrule
\end{tabular}
}

    \caption{Up to 20 exemplars generated by LLaMA-3.1-70B  (the best-performing model in terms of valid exemplars generated), sorted by availability score. We select the basic-level categories that produced the highest number of hallucinations, i.e., expressions unattested in the ItTenTen corpus. For the colouring rationale, see Table~\ref{tab:llama_exemplars}.}
    \label{tab:halls}
\end{table*}

\paragraph{Conceptual Confusion:} Cases in which the model misinterprets the intended sense or category of a lexical item, leading to the generation of exemplars that belong to a different semantic domain. For example, when prompted with \textit{margherita} as a flower (i.e., `daisy'), the model generates\textit{ d’Austria} (‘of Austria’), referencing Margherita d’Austria (Margaret of Parma, a historical figure\footnote{\url{https://en.wikipedia.org/wiki/Margaret_of_Parma}}), and \textit{d’Ungheria }(‘of Hungary’), referencing Margherita d’Ungheria (Saint Margaret of Hungary\footnote{\url{https://en.wikipedia.org/wiki/Margaret_of_Hungary_(saint)}}).

\paragraph{Imitation Based:} In this case, LLMs replicate the surface-level syntactic or morphological structure of a valid, attested exemplar, leading to the overgeneralization of that structure across subsequent, unattested or spurious exemplars. This imitation is often form-driven rather than grounded in semantic plausibility or real-world usage. This phenomenon typically arises when a salient exemplar introduces a productive or familiar template, which the model then extends combinatorially without regard for corpus evidence or conceptual appropriateness.
For instance, the attested exemplar TERRAZZO `alla veneziana` (Venetian PAVEMENT) serves as a template \texttt{NOUN + ADJECTIVE (ITALIAN LOCATION)} for generating further expressions like  \textit{terrazzo genovese, milanese, bergamasca, pavese, fiorentina}, none of which are attested or conventional within the category.
Similarly, for the concept CANDELABRO `a 5 bracci/a`, the syntactical structure `a N bracci/a' is reiterated multiple times with increasing numbers of arms.

\begin{table*}[!ht]
    \begin{subtable}[t]{\textwidth}
        \renewcommand*{\minval}{0.6}%
\renewcommand*{\maxval}{1.0}%

\resizebox{\linewidth}{!}{%

\begin{tabular}{lccccccccccccc}
\toprule
 & \MakeUppercase{animals} & \MakeUppercase{body parts} & \MakeUppercase{clothes} & \MakeUppercase{foods} & \MakeUppercase{furnishing} & \MakeUppercase{furniture} & \MakeUppercase{hobbies} & \MakeUppercase{housing} & \MakeUppercase{kitchen} & \MakeUppercase{plants} & \MakeUppercase{stationery} & \MakeUppercase{vehicles} & avg \\
\midrule
\texttt{llama-3.2-3B} & \gradient{0.76} & \gradient{0.81} & \gradient{0.82} & \gradient{0.67} & \gradient{0.95} & \gradient{0.92} & \gradient{0.87} & \gradient{0.73} & \gradient{0.83} & \gradient{0.94} & \gradient{0.94} & \gradient{0.8} & 0.84 \\
\texttt{llama-3.1-8B} & \gradient{1.0} & \gradient{0.94} & \gradient{1.0} & \gradient{0.93} & \gradient{1.0} & \gradient{0.92} & \gradient{0.93} & \gradient{0.93} & \gradient{0.92} & \gradient{1.0} & \gradient{1.0} & \gradient{1.0} & 0.96 \\
\texttt{llama-3.1-70B} & \gradient{1.0} & \gradient{0.94} & \gradient{0.94} & \gradient{1.0} & \gradient{1.0} & \gradient{0.92} & \gradient{0.93} & \gradient{0.93} & \gradient{0.92} & \gradient{0.94} & \gradient{1.0} & \gradient{0.93} & 0.95 \\
\texttt{mistral-7B} & \gradient{0.94} & \gradient{1.0} & \gradient{0.76} & \gradient{0.87} & \gradient{1.0} & \gradient{0.92} & \gradient{0.93} & \gradient{0.8} & \gradient{0.75} & \gradient{0.94} & \gradient{0.88} & \gradient{0.93} & 0.89 \\
\texttt{nemo-12B} & \gradient{0.94} & \gradient{1.0} & \gradient{1.0} & \gradient{1.0} & \gradient{1.0} & \gradient{0.83} & \gradient{1.0} & \gradient{1.0} & \gradient{0.92} & \gradient{0.88} & \gradient{0.94} & \gradient{0.93} & 0.95 \\
\texttt{mixtral-8x7B} & \gradient{0.94} & \gradient{1.0} & \gradient{0.94} & \gradient{1.0} & \gradient{1.0} & \gradient{1.0} & \gradient{1.0} & \gradient{1.0} & \gradient{1.0} & \gradient{0.94} & \gradient{1.0} & \gradient{1.0} & \textbf{0.98} \\
\midrule
\texttt{llava-7B} & \gradient{0.94} & \gradient{1.0} & \gradient{0.82} & \gradient{0.67} & \gradient{1.0} & \gradient{0.92} & \gradient{0.93} & \gradient{0.93} & \gradient{1.0} & \gradient{0.94} & \gradient{1.0} & \gradient{1.0} & 0.93 \\
\texttt{idefics2-8B} & \gradient{0.88} & \gradient{1.0} & \gradient{0.88} & \gradient{0.8} & \gradient{1.0} & \gradient{0.92} & \gradient{1.0} & \gradient{0.93} & \gradient{1.0} & \gradient{0.94} & \gradient{1.0} & \gradient{0.93} & 0.94 \\
\bottomrule
category avg & 0.93 & 0.96 & 0.90 & 0.87 & \textbf{0.99} & 0.92 & 0.95 & 0.91 & 0.92 & 0.94 & 0.97 & 0.94 & 0.93 \\
\bottomrule
\end{tabular}

}
\caption{Accuracy for \textbf{basic-level category} prediction.
}\label{tab:subtaskAbasic}

    \end{subtable}
    
    \vspace{0.25cm}
    
    \begin{subtable}[t]{\textwidth}
        \renewcommand*{\minval}{0.0}%
\renewcommand*{\maxval}{1.0}%

\resizebox{\linewidth}{!}{%

\begin{tabular}{lccccccccccccc}
\toprule
 & \MakeUppercase{animals} & \MakeUppercase{body parts} & \MakeUppercase{clothes} & \MakeUppercase{foods} & \MakeUppercase{furnishing} & \MakeUppercase{furniture} & \MakeUppercase{hobbies} & \MakeUppercase{housing} & \MakeUppercase{kitchen} & \MakeUppercase{plants} & \MakeUppercase{stationery} & \MakeUppercase{vehicles} & avg \\
\midrule
\texttt{llama-3.2-3B} & \gradient{0.94} & \gradient{0.12} & \gradient{0.71} & \gradient{0.07} & \gradient{0.0} & \gradient{0.75} & \gradient{0.07} & \gradient{0.8} & \gradient{1.0} & \gradient{0.81} & \gradient{0.0} & \gradient{0.93} & 0.52 \\
\texttt{llama-3.1-8B} & \gradient{1.0} & \gradient{0.81} & \gradient{0.76} & \gradient{0.2} & \gradient{0.0} & \gradient{0.92} & \gradient{0.13} & \gradient{0.8} & \gradient{1.0} & \gradient{0.94} & \gradient{0.0} & \gradient{1.0} & 0.63 \\
\texttt{llama-3.1-70B} & \gradient{1.0} & \gradient{0.69} & \gradient{0.35} & \gradient{0.4} & \gradient{0.0} & \gradient{1.0} & \gradient{0.07} & \gradient{0.93} & \gradient{1.0} & \gradient{0.88} & \gradient{0.44} & \gradient{0.93} & \textbf{0.64} \\
\texttt{mistral-7B} & \gradient{0.94} & \gradient{0.62} & \gradient{0.94} & \gradient{0.33} & \gradient{0.32} & \gradient{0.92} & \gradient{0.0} & \gradient{0.4} & \gradient{1.0} & \gradient{0.56} & \gradient{0.0} & \gradient{1.0} & 0.59 \\
\texttt{nemo-12B} & \gradient{0.06} & \gradient{0.81} & \gradient{0.12} & \gradient{0.0} & \gradient{0.0} & \gradient{1.0} & \gradient{0.07} & \gradient{0.2} & \gradient{1.0} & \gradient{0.75} & \gradient{0.5} & \gradient{1.0} & 0.46 \\
\texttt{mixtral-8x7B} & \gradient{1.0} & \gradient{0.94} & \gradient{0.06} & \gradient{0.47} & \gradient{0.0} & \gradient{0.83} & \gradient{0.13} & \gradient{0.6} & \gradient{1.0} & \gradient{0.75} & \gradient{0.06} & \gradient{1.0} & 0.57 \\
\midrule
\texttt{llava-7B} & \gradient{0.88} & \gradient{0.88} & \gradient{0.76} & \gradient{0.33} & \gradient{0.11} & \gradient{0.83} & \gradient{0.13} & \gradient{0.67} & \gradient{1.0} & \gradient{0.5} & \gradient{0.0} & \gradient{1.0} & 0.59 \\
\texttt{idefics2-8B} & \gradient{0.88} & \gradient{0.0} & \gradient{0.12} & \gradient{0.6} & \gradient{0.0} & \gradient{0.67} & \gradient{0.0} & \gradient{0.53} & \gradient{1.0} & \gradient{0.06} & \gradient{0.0} & \gradient{0.67} & 0.38 \\
\bottomrule
category avg & 0.84 & 0.61 & 0.48 & 0.30 & 0.05 & 0.86 & 0.08 & 0.62 & \textbf{1.00} & 0.66 & 0.12 & 0.94 & 0.53 \\
\bottomrule
\end{tabular}

}
\caption{Accuracy for \textbf{superordinate category} prediction.
}\label{tab:subtaskAsuper}

    \end{subtable}
\caption{\textsc{Subtask A}--Accuracy for category prediction at basic and super-ordinate category level.}
\end{table*}

\section{\textsc{Subtask A}}
\label{app:subtaskA}
In this section, we report the in-depth results for the experiment described in Section \ref{sec:super-preds}. Tables \ref{tab:subtaskAbasic} and \ref{tab:subtaskAsuper} report individual accuracy for each of the 12 superordinate categories for basic-level and superordinate-level category prediction, respectively.

\section{\textsc{Subtask B}}
\label{app:subtaskB}
In the following tables, we report individual accuracy for each of the 12 superordinate categories for \textsc{Subtask B}.
Results are grouped into three blocks according to the number of exemplars generated by the human subjects: (i) low coverage 
(up to 5 exemplars; Table \ref{tab:subtaskBlow}), (ii) medium coverage (6–10 exemplars; Table \ref{tab:subtaskBmedium}), and (iii) high coverage (more than 10 exemplars; Table \ref{tab:subtaskBhigh}). Note that the columns containing `na' values are the results of the frequency-based grouping. For example, we do not have any basic-level concept belonging to the super-ordinate category of \texttt{plants} that elicited a \textbf{high} number of exemplars in the human experimental phase. Hence, the empty column in Tables \ref{tab:subtaskBlow} and \ref{tab:subtaskBhigh}.

\begin{table*}
    \begin{subtable}[t]{\textwidth}
        \renewcommand*{\minval}{0.0}%
\renewcommand*{\maxval}{1.0}%

\resizebox{\linewidth}{!}{%

\begin{tabular}{lccccccccccccr}
\toprule
 & \MakeUppercase{animals} & \MakeUppercase{body parts} & \MakeUppercase{clothes} & \MakeUppercase{foods} & \MakeUppercase{furnishing} & \MakeUppercase{furniture} & \MakeUppercase{hobbies} & \MakeUppercase{housing} & \MakeUppercase{kitchen} & \MakeUppercase{plants} & \MakeUppercase{stationery} & \MakeUppercase{vehicles} & avg \\
\midrule
\texttt{llama-3.2-3B} & \gradient{0.38} & \gradient{0.60} & na & na & \gradient{0.60} & \gradient{0.50} & \gradient{0.67} & \gradient{0.75} & na & \gradient{0.77} & \gradient{1.00} & \gradient{0.60} & 0.65 \\
\texttt{llama-3.1-8B} & \gradient{0.38} & \gradient{0.80} & na & na & \gradient{0.60} & \gradient{0.25} & \gradient{0.33} & \gradient{0.75} & na & \gradient{0.54} & \gradient{1.00} & \gradient{0.60} & 0.58 \\
\texttt{llama-3.1-70B} & \gradient{0.38} & \gradient{1.00} & na & na & \gradient{0.80} & \gradient{0.75} & \gradient{0.67} & \gradient{0.75} & na & \gradient{0.85} & \gradient{1.00} & \gradient{0.40} & \textbf{0.73} \\
\texttt{mistral-7B} & \gradient{0.62} & \gradient{0.60} & na & na & \gradient{0.80} & \gradient{0.50} & \gradient{0.33} & \gradient{0.00} & na & \gradient{0.54} & \gradient{0.50} & \gradient{0.60} & 0.50 \\
\texttt{nemo-12B} & \gradient{0.25} & \gradient{0.40} & na & na & \gradient{0.60} & \gradient{0.50} & \gradient{0.67} & \gradient{0.75} & na & \gradient{0.69} & \gradient{0.50} & \gradient{0.40} & 0.53 \\
\texttt{mixtral-8x7B} & \gradient{0.50} & \gradient{1.00} & na & na & \gradient{0.80} & \gradient{0.50} & \gradient{0.67} & \gradient{1.00} & na & \gradient{0.69} & \gradient{0.50} & \gradient{0.80} & \textbf{0.72} \\
\midrule
\texttt{llava-7B} & \gradient{0.75} & \gradient{0.40} & na & na & \gradient{0.80} & \gradient{0.25} & \gradient{0.33} & \gradient{0.25} & na & \gradient{0.46} & \gradient{0.50} & \gradient{0.60} & 0.48 \\
\texttt{idefics2-8B} & \gradient{0.62} & \gradient{0.40} & na & na & \gradient{0.80} & \gradient{0.50} & \gradient{0.33} & \gradient{0.00} & na & \gradient{0.54} & \gradient{1.00} & \gradient{0.60} & 0.53 \\
\bottomrule
category avg & 0.48 & 0.65 & na & na & 0.72 & 0.47 & 0.50 & 0.53 & na & 0.63 & \textbf{0.75} & 0.57 & 0.59 \\
\bottomrule
\end{tabular}

} %
\caption{\textbf{Low coverage} basic-level categories.}\label{tab:subtaskBlow}

    \end{subtable}
    
    \vspace{0.25cm}

    \begin{subtable}[t]{\textwidth}
        \renewcommand*{\minval}{0.0}%
\renewcommand*{\maxval}{1.0}%

\resizebox{\linewidth}{!}{%

\begin{tabular}{lccccccccccccr}
\toprule
 & \MakeUppercase{animals} & \MakeUppercase{body parts} & \MakeUppercase{clothes} & \MakeUppercase{foods} & \MakeUppercase{furnishing} & \MakeUppercase{furniture} & \MakeUppercase{hobbies} & \MakeUppercase{housing} & \MakeUppercase{kitchen} & \MakeUppercase{plants} & \MakeUppercase{stationery} & \MakeUppercase{vehicles} & avg \\
 \midrule
\texttt{llama-3.2-3B} & \gradient{0.50} & \gradient{0.33} & \gradient{1.00} & \gradient{0.00} & \gradient{0.67} & \gradient{1.00} & \gradient{0.75} & \gradient{0.67} & \gradient{0.67} & \gradient{0.80} & \gradient{0.67} & \gradient{0.33} & 0.62 \\
\texttt{llama-3.1-8B} & \gradient{0.75} & \gradient{0.44} & \gradient{0.80} & \gradient{0.50} & \gradient{0.44} & \gradient{0.60} & \gradient{0.75} & \gradient{0.83} & \gradient{0.33} & \gradient{0.60} & \gradient{0.78} & \gradient{0.33} & 0.60 \\
\texttt{llama-3.1-70B} & \gradient{0.75} & \gradient{0.33} & \gradient{1.00} & \gradient{1.00} & \gradient{0.67} & \gradient{0.60} & \gradient{0.75} & \gradient{0.67} & \gradient{0.33} & \gradient{0.80} & \gradient{0.78} & \gradient{0.50} & \textbf{0.68} \\
\texttt{mistral-7B} & \gradient{0.50} & \gradient{0.44} & \gradient{0.80} & \gradient{1.00} & \gradient{0.56} & \gradient{0.60} & \gradient{0.75} & \gradient{0.67} & \gradient{0.50} & \gradient{0.20} & \gradient{0.33} & \gradient{0.50} & 0.57 \\
\texttt{nemo-12B} & \gradient{0.75} & \gradient{0.56} & \gradient{0.80} & \gradient{1.00} & \gradient{0.67} & \gradient{1.00} & \gradient{0.75} & \gradient{0.50} & \gradient{0.67} & \gradient{0.80} & \gradient{0.44} & \gradient{0.33} & \textbf{0.69} \\
\texttt{mixtral-8x7B} & \gradient{0.75} & \gradient{0.56} & \gradient{1.00} & \gradient{0.50} & \gradient{0.67} & \gradient{0.40} & \gradient{0.75} & \gradient{0.33} & \gradient{0.67} & \gradient{0.20} & \gradient{0.33} & \gradient{0.50} & 0.55 \\
\midrule
\texttt{llava-7B} & \gradient{0.25} & \gradient{0.33} & \gradient{0.80} & \gradient{1.00} & \gradient{0.56} & \gradient{0.80} & \gradient{0.75} & \gradient{0.67} & \gradient{0.33} & \gradient{0.80} & \gradient{0.44} & \gradient{0.67} & 0.62 \\
\texttt{idefics2-8B} & \gradient{0.25} & \gradient{0.22} & \gradient{0.80} & \gradient{1.00} & \gradient{0.67} & \gradient{0.80} & \gradient{0.75} & \gradient{0.50} & \gradient{0.17} & \gradient{0.80} & \gradient{0.44} & \gradient{0.50} & 0.58 \\
\bottomrule
category avg & 0.56 & 0.40 & \textbf{0.88} & 0.75 & 0.61 & 0.72 & 0.75 & 0.60 & 0.46 & 0.62 & 0.53 & 0.46 & 0.61 \\
\bottomrule
\end{tabular}

}

\caption{\textbf{Medium coverage} basic-level categories.}\label{tab:subtaskBmedium}

    \end{subtable}
    
    \vspace{0.25cm}
    
    \begin{subtable}[t]{\textwidth}
        \renewcommand*{\minval}{0.0}%
\renewcommand*{\maxval}{1.0}%

\resizebox{\linewidth}{!}{%

\begin{tabular}{lccccccccccccr}
\toprule
 & \MakeUppercase{animals} & \MakeUppercase{body parts} & \MakeUppercase{clothes} & \MakeUppercase{foods} & \MakeUppercase{furnishing} & \MakeUppercase{furniture} & \MakeUppercase{hobbies} & \MakeUppercase{housing} & \MakeUppercase{kitchen} & \MakeUppercase{plants} & \MakeUppercase{stationery} & \MakeUppercase{vehicles} & avg \\

\midrule
\texttt{llama-3.2-3B} & \gradient{0.60} & \gradient{0.50} & \gradient{0.58} & \gradient{0.62} & \gradient{0.20} & \gradient{0.33} & \gradient{0.38} & \gradient{0.40} & \gradient{0.17} & na & \gradient{0.40} & \gradient{0.50} & 0.42 \\
\texttt{llama-3.1-8B} & \gradient{0.60} & \gradient{0.00} & \gradient{0.42} & \gradient{0.54} & \gradient{0.60} & \gradient{0.00} & \gradient{0.50} & \gradient{0.40} & \gradient{0.50} & na & \gradient{0.60} & \gradient{0.50} & 0.42 \\
\texttt{llama-3.1-70B} & \gradient{0.40} & \gradient{1.00} & \gradient{0.83} & \gradient{0.77} & \gradient{0.60} & \gradient{0.67} & \gradient{0.50} & \gradient{0.40} & \gradient{0.67} & na & \gradient{0.40} & \gradient{0.50} & \textbf{0.61 }\\
\texttt{mistral-7B} & \gradient{0.20} & \gradient{0.50} & \gradient{0.50} & \gradient{0.62} & \gradient{0.40} & \gradient{0.33} & \gradient{0.38} & \gradient{0.40} & \gradient{0.67} & na & \gradient{0.40} & \gradient{0.75} & 0.47 \\
\texttt{nemo-12B} & \gradient{0.60} & \gradient{1.00} & \gradient{0.33} & \gradient{0.69} & \gradient{0.80} & \gradient{0.33} & \gradient{0.12} & \gradient{0.40} & \gradient{0.50} & na & \gradient{0.40} & \gradient{0.50} & 0.52 \\
\texttt{mixtral-8x7B} & \gradient{0.80} & \gradient{0.00} & \gradient{0.50} & \gradient{0.62} & \gradient{0.80} & \gradient{0.67} & \gradient{0.25} & \gradient{0.20} & \gradient{0.67} & na & \gradient{0.80} & \gradient{1.00} & 0.57 \\
\midrule
\texttt{llava-7B} & \gradient{0.60} & \gradient{0.50} & \gradient{0.42} & \gradient{0.77} & \gradient{0.40} & \gradient{0.33} & \gradient{0.25} & \gradient{0.40} & \gradient{0.50} & na & \gradient{0.40} & \gradient{0.75} & 0.48 \\
\texttt{idefics2-8B} & \gradient{0.80} & \gradient{0.50} & \gradient{0.42} & \gradient{0.69} & \gradient{0.40} & \gradient{0.00} & \gradient{0.25} & \gradient{0.40} & \gradient{0.33} & na & \gradient{0.40} & \gradient{0.75} & 0.45 \\
\bottomrule
category avg & 0.57 & 0.50 & 0.50 & \textbf{0.66} & 0.52 & 0.33 & 0.33 & 0.38 & 0.50 & na & 0.48 & \textbf{0.66} & 0.49 \\
\bottomrule

\end{tabular}

}
\caption{\textbf{High coverage} basic-level categories.}\label{tab:subtaskBhigh}

    \end{subtable}
\caption{\textsc{Subtask B}--Typicality Accuracy at \textbf{different coverage} of basic-level categories.}
\end{table*}

\subsection{SUBTASK B: Typicality Variation by Availability Score}
In this Section, we report the results for the typicality prediction experiments described in Section \ref{sec:typicality} by aggregating the results along the availability score. Specifically, we group results according to the absolute difference between the availability score of the most-available exemplars and the availability score of the least-available one. The availability score is computed on the human experiment's results. 

\begin{table*}
    \begin{subtable}[t]{\textwidth}
        \renewcommand*{\minval}{0.0}%
\renewcommand*{\maxval}{1.0}%

\resizebox{\linewidth}{!}{%

\begin{tabular}{lccccccccccccr}
\toprule
 & \MakeUppercase{animals} & \MakeUppercase{body parts} & \MakeUppercase{clothes} & \MakeUppercase{foods} & \MakeUppercase{furnishing} & \MakeUppercase{furniture} & \MakeUppercase{hobbies} & \MakeUppercase{housing} & \MakeUppercase{kitchen} & \MakeUppercase{plants} & \MakeUppercase{stationery} & \MakeUppercase{vehicles} & avg \\
 
\midrule
\texttt{llama-3.2-3B} & \gradient{0.50} & \gradient{0.33} & \gradient{0.75} & \gradient{0.71} & \gradient{0.80} & \gradient{1.00} & \gradient{0.57} & \gradient{0.67} & \gradient{0.38} & \gradient{1.00} & \gradient{0.60} & \gradient{0.00} & 0.61 \\
\texttt{llama-3.1-8B} & \gradient{0.75} & \gradient{0.33} & \gradient{0.75} & \gradient{0.71} & \gradient{0.80} & \gradient{0.50} & \gradient{0.57} & \gradient{0.67} & \gradient{0.50} & \gradient{0.50} & \gradient{0.80} & \gradient{0.00} & 0.57 \\
\texttt{llama-3.1-70B} & \gradient{0.25} & \gradient{0.67} & \gradient{1.00} & \gradient{1.00} & \gradient{1.00} & \gradient{1.00} & \gradient{0.71} & \gradient{0.67} & \gradient{0.50} & \gradient{1.00} & \gradient{0.60} & \gradient{0.00} & \textbf{0.70} \\
\texttt{mistral-7B} & \gradient{0.25} & \gradient{0.67} & \gradient{0.50} & \gradient{0.86} & \gradient{0.60} & \gradient{1.00} & \gradient{0.71} & \gradient{0.50} & \gradient{0.62} & \gradient{0.00} & \gradient{0.60} & \gradient{0.00} & 0.53 \\
\texttt{nemo-12B} & \gradient{0.50} & \gradient{1.00} & \gradient{0.75} & \gradient{0.71} & \gradient{1.00} & \gradient{1.00} & \gradient{0.71} & \gradient{0.67} & \gradient{0.75} & \gradient{1.00} & \gradient{0.20} & \gradient{0.00} & 0.69 \\
\texttt{mixtral-8x7B} & \gradient{0.75} & \gradient{0.33} & \gradient{0.75} & \gradient{0.86} & \gradient{1.00} & \gradient{1.00} & \gradient{0.71} & \gradient{0.50} & \gradient{0.62} & \gradient{0.50} & \gradient{0.80} & \gradient{0.50} & 0.69 \\
\midrule
\texttt{llava-7B} & \gradient{0.50} & \gradient{0.33} & \gradient{0.50} & \gradient{0.86} & \gradient{0.80} & \gradient{1.00} & \gradient{0.57} & \gradient{0.67} & \gradient{0.50} & \gradient{0.50} & \gradient{0.60} & \gradient{0.50} & 0.61 \\
\texttt{idefics2-8B} & \gradient{0.50} & \gradient{0.67} & \gradient{0.50} & \gradient{0.71} & \gradient{0.80} & \gradient{0.50} & \gradient{0.57} & \gradient{0.33} & \gradient{0.25} & \gradient{0.50} & \gradient{0.60} & \gradient{0.00} & 0.49 \\
\midrule
category avg & 0.50 & 0.54 & 0.69 & 0.80 & 0.85 & \textbf{0.88} & 0.64 & 0.58 & 0.52 & 0.62 & 0.60 & 0.12 & 0.61 \\
\bottomrule
\end{tabular}}
\caption{\textbf{High absolute difference} in availability score ($ |\Delta| > 0.4$).}

    \end{subtable}

    \vspace{0.25cm}

    \begin{subtable}[t]{\textwidth}
        \renewcommand*{\minval}{0.0}%
\renewcommand*{\maxval}{1.0}%

\resizebox{\linewidth}{!}{%

\begin{tabular}{lccccccccccccr}
\toprule
 & \MakeUppercase{animals} & \MakeUppercase{body parts} & \MakeUppercase{clothes} & \MakeUppercase{foods} & \MakeUppercase{furnishing} & \MakeUppercase{furniture} & \MakeUppercase{hobbies} & \MakeUppercase{housing} & \MakeUppercase{kitchen} & \MakeUppercase{plants} & \MakeUppercase{stationery} & \MakeUppercase{vehicles} & avg \\

\midrule
\texttt{llama-3.2-3B} & \gradient{0.50} & \gradient{0.60} & \gradient{0.86} & \gradient{0.38} & \gradient{0.38} & \gradient{0.60} & \gradient{0.40} & \gradient{0.80} & \gradient{0.50} & \gradient{0.67} & \gradient{0.71} & \gradient{0.57} & 0.58 \\
\texttt{llama-3.1-8B} & \gradient{0.50} & \gradient{0.60} & \gradient{0.57} & \gradient{0.38} & \gradient{0.50} & \gradient{0.20} & \gradient{0.60} & \gradient{0.80} & \gradient{0.25} & \gradient{0.33} & \gradient{0.71} & \gradient{0.57} & 0.50 \\
\texttt{llama-3.1-70B} & \gradient{0.50} & \gradient{0.70} & \gradient{0.86} & \gradient{0.62} & \gradient{0.62} & \gradient{0.40} & \gradient{0.60} & \gradient{0.60} & \gradient{0.50} & \gradient{0.67} & \gradient{0.71} & \gradient{0.57} & \textbf{0.61} \\
\texttt{mistral-7B} & \gradient{0.50} & \gradient{0.50} & \gradient{0.57} & \gradient{0.50} & \gradient{0.62} & \gradient{0.40} & \gradient{0.40} & \gradient{0.40} & \gradient{0.50} & \gradient{0.33} & \gradient{0.29} & \gradient{0.86} & 0.49 \\
\texttt{nemo-12B} & \gradient{0.50} & \gradient{0.50} & \gradient{0.29} & \gradient{0.75} & \gradient{0.62} & \gradient{0.60} & \gradient{0.20} & \gradient{0.40} & \gradient{0.25} & \gradient{0.67} & \gradient{0.57} & \gradient{0.57} & 0.49 \\
\texttt{mixtral-8x7B} & \gradient{0.50} & \gradient{0.70} & \gradient{0.71} & \gradient{0.38} & \gradient{0.88} & \gradient{0.40} & \gradient{0.40} & \gradient{0.40} & \gradient{0.75} & \gradient{0.33} & \gradient{0.29} & \gradient{0.86} & 0.55 \\
\midrule
\texttt{llava-7B} & \gradient{0.50} & \gradient{0.50} & \gradient{0.29} & \gradient{0.75} & \gradient{0.50} & \gradient{0.60} & \gradient{0.20} & \gradient{0.40} & \gradient{0.25} & \gradient{0.33} & \gradient{0.43} & \gradient{0.86} & 0.47 \\
\texttt{idefics2-8B} & \gradient{0.50} & \gradient{0.30} & \gradient{0.43} & \gradient{0.75} & \gradient{0.75} & \gradient{0.40} & \gradient{0.20} & \gradient{0.40} & \gradient{0.25} & \gradient{0.33} & \gradient{0.43} & \gradient{0.86} & 0.47 \\
\midrule
category avg & 0.50 & 0.55 & 0.57 & 0.56 & 0.61 & 0.45 & 0.38 & 0.52 & 0.41 & 0.46 & 0.52 & \textbf{0.71} & 0.52 \\
\bottomrule
\end{tabular}}
\caption{\textbf{Medium absolute difference} in availability score ($0.2 \leq |\Delta| \leq 0.4$).}

    \end{subtable}

    \vspace{0.5cm}

    \begin{subtable}[t]{\textwidth}
        \renewcommand*{\minval}{0.0}%
\renewcommand*{\maxval}{1.0}%

\resizebox{\linewidth}{!}{%

\begin{tabular}{lccccccccccccr}
\toprule
 & \MakeUppercase{animals} & \MakeUppercase{body parts} & \MakeUppercase{clothes} & \MakeUppercase{foods} & \MakeUppercase{furnishing} & \MakeUppercase{furniture} & \MakeUppercase{hobbies} & \MakeUppercase{housing} & \MakeUppercase{kitchen} & \MakeUppercase{plants} & \MakeUppercase{stationery} & \MakeUppercase{vehicles} & avg \\

\midrule
\texttt{llama-3.2-3B} & \gradient{0.43} & \gradient{0.00} & \gradient{0.50} &na & \gradient{0.50} & \gradient{0.60} & \gradient{0.67} & \gradient{0.25} & na &\gradient{0.73} & \gradient{0.50} & \gradient{0.50} & 0.47 \\
\texttt{llama-3.1-8B} & \gradient{0.43} & \gradient{0.33} & \gradient{0.33} &na & \gradient{0.33} & \gradient{0.40} & \gradient{0.33} & \gradient{0.50} &na & \gradient{0.64} & \gradient{0.75} & \gradient{0.50} & 0.45 \\
\texttt{llama-3.1-70B} & \gradient{0.57} & \gradient{0.33} & \gradient{0.83} & na &\gradient{0.50} & \gradient{0.80} & \gradient{0.33} & \gradient{0.50} &na & \gradient{0.82} & \gradient{0.75} & \gradient{0.50} & \textbf{0.59} \\
\texttt{mistral-7B} & \gradient{0.57} & \gradient{0.33} & \gradient{0.67} &na & \gradient{0.50} & \gradient{0.40} & \gradient{0.00} & \gradient{0.25} & na &\gradient{0.64} & \gradient{0.25} & \gradient{0.50} & 0.41 \\
\texttt{nemo-12B} & \gradient{0.43} & \gradient{0.33} & \gradient{0.50} & na &\gradient{0.50} & \gradient{0.60} & \gradient{0.00} & \gradient{0.50} & na &\gradient{0.64} & \gradient{0.50} & \gradient{0.33} & 0.43 \\
\texttt{mixtral-8x7B} & \gradient{0.71} & \gradient{0.67} & \gradient{0.50} &na & \gradient{0.33} & \gradient{0.40} & \gradient{0.00} & \gradient{0.50} & na &\gradient{0.64} & \gradient{0.50} & \gradient{0.67} & 0.49 \\
\midrule
\texttt{llava-7B} & \gradient{0.71} & \gradient{0.00} & \gradient{0.83} & na &\gradient{0.50} & \gradient{0.20} & \gradient{0.33} & \gradient{0.25} & na &\gradient{0.64} & \gradient{0.25} & \gradient{0.50} & 0.42 \\
\texttt{idefics2-8B} & \gradient{0.71} & \gradient{0.00} & \gradient{0.67} & na &\gradient{0.33} & \gradient{0.60} & \gradient{0.33} & \gradient{0.25} & na &\gradient{0.73} & \gradient{0.50} & \gradient{0.50} & 0.46 \\
\midrule
category avg & \textbf{0.57} & 0.25 & 0.60 & na &0.44 & 0.50 & 0.25 & 0.38 & na &0.68 & 0.50 & 0.50 & 0.47 \\

\bottomrule
\end{tabular}}
\caption{\textbf{Low absolute difference} in availability score ($|\Delta| < 0.2$).}

    \end{subtable}
\caption{\textsc{Subtask B}--Typicality Accuracy at \textbf{different availability score} of exemplars.}
\end{table*}

\end{document}